\documentclass[10pt,twocolumn,letterpaper]{article}

\usepackage{cvpr}             
\usepackage{algorithmic}
\usepackage{multirow}

\usepackage{float}
\usepackage{tabularx}
\usepackage{array}

\usepackage{xcolor}

\definecolor{cvprblue}{rgb}{0.21,0.49,0.74}
\usepackage[pagebackref,breaklinks,colorlinks,allcolors=cvprblue]{hyperref}

\usepackage{orcidlink}

\def\paperID{30}
\def\confName{CVPR}
\def\confYear{2026}

\title{City-Mesh3R: Simulation-Ready City-Scale 3D Mesh Reconstruction \\ from Multi-View Images}

\author{Sayan Paul \orcidlink{0000-0001-9885-233X}, Sourav Ghosh, Siddharth Katageri, Soumyadip Maity, Sanjana Sinha, Brojeshwar Bhowmick\\
{\small \{p.sayan, g.sourav10, siddharth.katageri, soumyadip.maity, sanjana.sinha, b.bhowmick\} $@$ tcs.com} \\
{\small \textbf{Visual Computing \& Embodied AI Lab, TCS Research, India}}
}

\makeatletter
\apptocmd{\@maketitle}{
  \begin{center}
    \includegraphics[width=\textwidth]{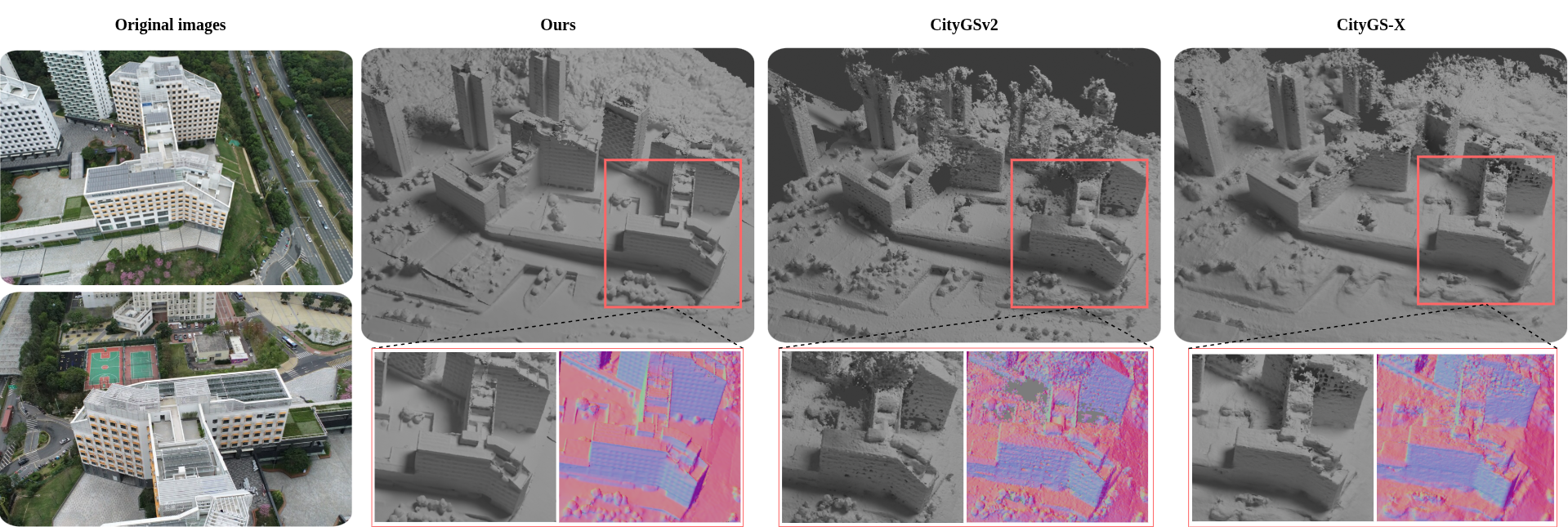}
    \captionof{figure}{Our method reconstructs simulation-ready watertight 3D meshes with high geometric fidelity, smooth surface normals, from input multi-view images of large-scale urban scenes from GauU-Scene \cite{xiong2024gauuscenescenereconstructionbenchmark} dataset. Existing methods \cite{gao2025citygs,liu2024citygaussianv2} reconstruct 3D meshes with irregular surface normals, missing regions, inaccurate surface geometry details.\\}
    \label{teaser}
  \end{center}
}{}{}
\makeatother

\begin{document}
\maketitle

\begin{abstract}
City-scale 3D surface reconstruction from multiview images for downstream 3D simulation, poses highly challenging problems due to the scale and complexity of urban scenes. Existing city-scale 3D reconstruction methods based on NeRF, Gaussian Splatting etc. often fail to recover 3D meshes ready for simulation due to incomplete/missing geometry and irregular, noisy surfaces. Scaling existing small-scale 3D reconstruction methods to arbitrarily large urban scenes is highly infeasible due to their computational complexity. We present City-Mesh3R, a scalable framework for reconstructing watertight surface meshes directly from large unordered image collections. Unlike recent methods which use global sparse SfM point-cloud initialization followed by a distributed 3D dense reconstruction of large-scale scenes, our method follows an end-to-end images-to-mesh 3D reconstruction approach using a divide-and-conquer strategy. The sparse city map is reconstructed via topological image clustering, cluster-wise independent sparse SfM and map merging, without need for exhaustive image feature matching. Then this map is partitioned spatially to perform geometry-aware camera selection, followed by dense surface reconstruction and surface refinement using curvature-aware adaptive vertex density remeshing. These partition meshes are then stitched together to produce the global mesh of the city. The proposed end-to-end framework is evaluated on city-scale reconstruction datasets. As demonstrated by our qualitative and quantitative results, our proposed method yields high-fidelity watertight 3D meshes with regular geometry, capturing fine surface details, and is suitable for scaling to arbitrarily large scenes owing to the end-to-end processing in a distributed setting.
\end{abstract}
    
\section{Introduction}{
\label{sec:intro}

3D reconstruction from multiview images of city-scale scenes \cite{City10892091} is a challenging research problem which directly impacts applications such as urban digital twin \cite{3dcitymodel}, 15-minute city planning\cite{iqbal2025exploring}, infrastructure monitoring and disaster management. These applications often require simulation of high-fidelity city-scale 3D meshes that can capture the structural details of buildings, public infrastructure etc.  in real-world urban scenes.  However, scaling to arbitrarily large city scenes is challenging owing to a trade-off between geometric fidelity and computational complexity. 

Recent research use NeRF \cite{tancik2022block, turki2022mega,zhang2024supernerf} or Gaussian Splatting representations \cite{lin2024vastgaussian,liu2024citygaussian,liu2024citygaussianv2} for city-scale 3D reconstruction. 3D Surface reconstruction using Gaussian Splat representations can result in ill-defined surface boundaries, making it challenging to extract consistent normals, preserve sharp features, or perform structural edits such as adding or reshaping buildings. For downstream tasks like for e.g. urban planning, infrastructure modification, and hydrodynamic flood simulation; explicit mesh-based representations remain the practical and simulation-ready solution. Extracting 3D mesh surfaces from NeRF or Gaussian Splats often results in noisy, incomplete surfaces (holes or tears) that need manual postprocessing for simulation. Moreover, NeRF-based methods are computationally slow, while Gaussian-Splat–based approaches incur additional overhead for mesh extraction. Inaccuracy in 3D geometry directly impacts applications such as urban infrastructure monitoring, disaster management, etc. Moreover, methods based on gaussian-splatting rely on a good SfM point cloud initialization \cite{schoenberger2016sfm} typically computed using exhaustive image feature matching with huge computational complexity. Also, existing methods for 3D mesh reconstruction cannot scale to arbitrarily large city scenes captured by thousands or millions of images.  

In this work, we present a novel pipeline for scalable 3D mesh reconstruction that produces simulation-ready watertight 3D meshes of city-scale scenes. Following large-scale 3D reconstruction methods, we apply a distributed approach to partition the 3D urban scene, reconstruct individual partitions and merge the individual maps into a global 3D city map. Our proposed pipeline applies the distributed reconstruction in a sparse-to-dense manner. It consists of a distributed SfM stage to generate a city-scale sparse point cloud followed by a distributed approach for 3D mesh reconstruction and surface optimization. Unlike the recent city-scale mesh reconstruction techniques that use a given initial sparse global point cloud from a non-scalable standard SfM pipeline \cite{schoenberger2016sfm}, our proposed end-to-end pipeline reconstructs both sparse point cloud and 3D mesh in a distributed manner, making it easy to scale to arbitrarily large scenes.  Unlike city-scale reconstruction methods using NeRF/GS which perform global 3D mesh surface extraction and are not scalable, our proposed method follows a distributed approach for 3D mesh reconstruction and optimization, followed by mesh stitching. 

Instead of relying on exhaustive pairwise feature matching in a standard SfM pipeline \cite{schoenberger2016sfm,pan2024glomap}, we apply community-based image clustering to obtain  coarse topological clusters of the scene for distributed sparse SfM. The sparse point clouds reconstructed for overlapping clusters are merged to reconstruct a globally aligned sparse map of the entire city. We then apply geometry-aware partitioning of the globally aligned sparse map, leveraging camera poses and co-visibility statistics to produce data partitions suitable for downstream dense reconstruction. This hierarchical partitioning strategy yields well-formed map partitions, mitigates redundancy and ambiguity caused by irregular topological cluster boundaries, and significantly improves the robustness of subsequent dense map merging. We adapt a fast mesh optimization and remeshing procedure based on differentiable rendering for each partition. Our formulation introduces adaptive vertex density remeshing based on local scene complexity instead of uniform distribution, to produce watertight meshes with fine structural details. This adaptive formulation facilitates accurate geometric recovery, better optimization stability and memory efficiency, making it particularly well suited for large-scale urban environments. \\

\noindent
\textbf{Our main contributions} are as follows:
\begin{itemize}

    \item We present a distributed hierarchical sparse-to-dense pipeline for reconstructing city-scale, simulation-ready, high-fidelity watertight surface meshes from large unordered image collections, enabling scalability to arbitrarily large scenes.
    \item We propose a two-stage divide-and-conquer strategy for large-scale reconstruction that decouples sparse and dense partitioning to match their different data requirements: topological image clustering for distributed SfM, followed by spatial partitioning with geometry-aware camera selection for dense reconstruction.    
    \item We introduce a curvature-aware adaptive remeshing strategy for large urban scenes that allocates more vertices to geometrically complex regions and fewer to flatter areas, improving reconstruction quality, stability, and memory efficiency.
    \item We establish the efficacy of our proposed pipeline in reduced computational complexity and improved mesh fidelity on urban-scale 3D benchmark datasets.

\end{itemize}

}

\section{Related Work}

Large-scale 3D reconstruction has been explored through diverse representations, including point clouds, volumetric NeRFs, Gaussian splats, and meshes. Early systems such as \cite{agarwal2011building} demonstrated city-scale sparse point cloud reconstruction from 150,000 images via distributed matching and parallel SfM, while large-scale Structure-from-Motion \cite{agarwal2011rome, schoenberger2016sfm} and MVS methods \cite{orsingher2022revisiting} methods do not focus on reconstruction mesh surfaces.  \cite{zhu2018very} introduced the first million-image global SfM using a divide-and-conquer camera partitioning framework.

Recent research on large-scale 3D reconstruction increasingly adopts neural scene representations. NeRF-based approaches \cite{tancik2022block,turki2022mega,zhang2024supernerf} perform block-wise training for large scenes, and SDF-based models \cite{yang2025scalable} leverage dense fields for geometry extraction. Recent methods apply Gaussian-splatting \cite{lin2024vastgaussian,kerbl2024hierarchical,chen2024dogs,liu2024citygaussian,liu2024citygaussianv2,chen2025gigags,xiong2024sa} to enable real-time rendering of large-scale urban scenes, with CityGaussian \cite{liu2024citygaussian} and CityGaussianV2 \cite{liu2024citygaussianv2} scaling 3DGS training to multi-square-kilometer scenes; CityGS-X \cite{gao2025citygs} accelerates this using multitask rendering; and DOGS \cite{chen2024dogs} applies ADMM-based block optimization. Horizon-GS \cite{jiang2025horizon} fuses aerial and street views using 3DGS for rendering and 2DGS for geometry.

Surface extraction from Gaussian or neural fields has been explored through Poisson surface reconstruction in SuGaR \cite{guedon2024sugar}, SDF-based depth fusion in 2DGS \cite{huang20242d}, opacity-field iso-surfacing in GOF \cite{yu2024gaussian}, planar-disk splatting in PGSR \cite{chen2024pgsr}, and hybrid SDF–GS methods such as 3DGSR \cite{lyu20243dgsr} and SurfaceSplat \cite{gao2025surfacesplat}. Additional formulations include sparse-voxel rasterization (GeoSVR \cite{li2025geosvr,sun2025sparse}) and differentiable mesh extraction in MILo \cite{guedon2025milo}. MeshSplatting \cite{Held2025MeshSplatting} offers an efficient representation but doesn't guarantee watertight mesh or manifold-ness, while RadianceMeshes \cite{mai2025radiancemeshesvolumetricreconstruction} produces meshes at the cost of significantly longer reconstruction time. However, both methods remain unsuitable for distributed city-scale reconstruction due to their high per-block computational overhead. Large-scale mesh methods \cite{chen2025gigags,jiang2025horizon,gao2025citygs} typically rely on global TSDF fusion, which is computationally expensive, fails to capture thin structures, and often produces irregular surfaces with holes \cite{liu2024citygaussianv2}.

Moreover, Gaussian-based surface extraction often suffers from ill-defined boundaries, noisy geometry, and difficulty preserving sharp edges, limiting its applicability to downstream tasks such as urban planning, infrastructure modeling, or flood simulation. Similarly, mesh extraction from NeRF-based methods often produces irregular or incomplete surfaces and incurs substantial computation time, making these approaches impractical for large-scale reconstruction. In contrast, our method is explicitly designed for practical scalability and produces simulation-ready mesh outputs, making it directly suitable for real-world city-scale digital-twin applications.

\begin{figure*}[t] 
    \centering
    \includegraphics[width=\textwidth]{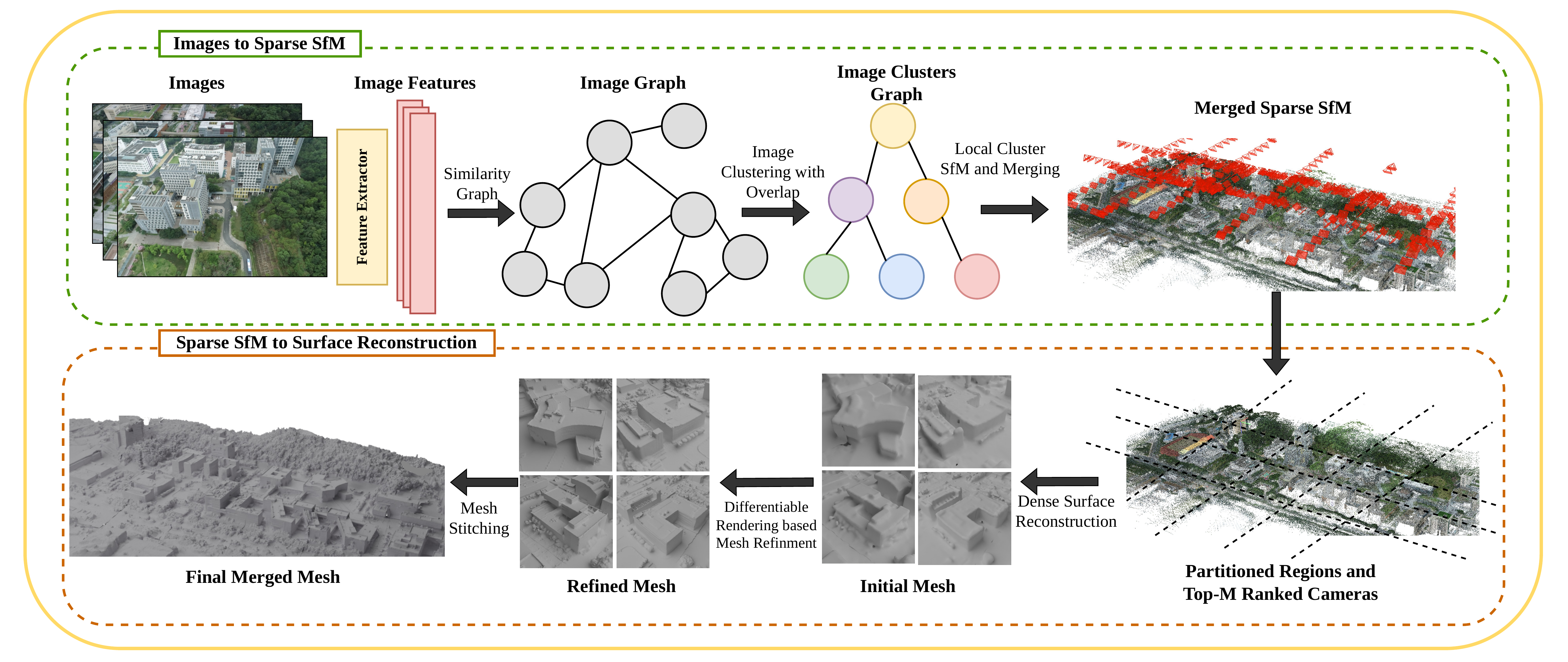}
    \caption{Our proposed pipeline: Starting from an unordered input image set, we first build a sparse SfM representation of the entire scene by retrieving related images with global descriptors, organizing them into overlapping clusters, reconstructing each cluster locally, and then hierarchically aligning and merging the resulting partial reconstructions into a unified sparse model. Given the sparse reconstruction, we then partition the scene spatially and, for each partition, select a compact yet informative subset of cameras for local dense processing. Using these fixed SfM cameras together with depth predictions from zero-shot foundational models, we construct an aligned surface initialization through depth alignment, multi-view consistency refinement, and volumetric fusion. Then, we optimize the reconstructed mesh using differential rendering and curvature-aware adaptive vertex density remeshing scheme. Finally, we merge neighboring partition meshes into a single coherent mesh surface.}

    \label{fig01_pipeline_arch_diag}
\end{figure*}

\section{Methodology}

Given a sequence of unordered images  $ \mathcal{I} = \{I_1, I_2, \dots, I_N\}$ capturing a large-scale scene, our proposed method (shown in Fig. \ref{fig01_pipeline_arch_diag}) reconstructs the global mesh representing the scene. Our proposed pipeline first reconstructs a sparse 3D point cloud of the entire scene using a distributed SfM approach (Images to Sparse SFM stage). Then the sparse 3D point cloud is converted to a watertight mesh via dense surface reconstruction followed by mesh surface refinement in a partitioned approach (Sparse SFM to Surface Reconstruction stage). The next sections describe the different stages of our proposed pipeline.

\subsection{Images to Sparse SfM}

\textbf{\textit{Global Feature Extraction and Similarity Graph Construction: }} Traditional large-scale SfM systems (e.g., \cite{chen2020graphbasedparallellargescale}) rely on local feature matching (e.g., SIFT) to construct the image view graph, requiring exhaustive pairwise descriptor matching, which becomes computationally prohibitive at scale. To address this, modern pipelines employ global image descriptors such as AP-GeM~\cite{revaud2019learningaverageprecisiontraining} and distilled NetVLAD~\cite{arandjelovic2016netvladcnnarchitectureweakly} used in ~\cite{humenberger2022robustimageretrievalbasedvisual} and ~\cite{sarlin2019coarsefinerobusthierarchical} for efficient image retrieval. These approaches adopt a two-stage strategy: global descriptors first identify candidate image pairs, followed by local feature matching, significantly reducing computational cost and improving scalability.
We construct the view graph using global
image descriptors. Specifically, we employ the DINOv2 \cite{oquab2024dinov2learningrobustvisual} vision transformer
model to extract global features from images due to its strong
generalization capability across diverse datasets.

For each image $I_i$, we compute a global feature vector
$ \mathbf{f} = \Phi(I) \in \mathbb{R}^d $,
where $\Phi(\cdot)$ denotes the DINOv2 feature extractor.
Using these feature vectors, we construct an image similarity graph  $G = (V, E)$, 
where each vertex $v_i \in V$ corresponds to an image $I_i$.
An edge $(i,j) \in E$ is created if the similarity score satisfies
$ d_S(\mathbf{f}_i,\mathbf{f}_j) \geq \tau $
,where $d_S(\cdot,\cdot)$ denotes the cosine similarity value and
$\tau$ is a predefined similarity threshold. 

\textbf{\textit{Image clustering with overlaps: }}
\label{OverLapIC} Commonly used clustering approaches such as Normalized Cuts
(Ncut) in \cite{chen2020graphbasedparallellargescale} or Agglomerative Clustering \cite{scikit-learn} typically produce disjoint clusters
and do not guarantee overlaps between them. However, for our distributed sparse SfM we need some overlap between clusters for a reliable merging into a global 3D map. To achieve this, we apply the Speaker–Listener Label Propagation Algorithm (SLPA) \cite{xie2011slpa},
which naturally detects overlapping communities in graphs, thereby
producing clusters that share common images and facilitate subsequent merging of individually reconstructed point clouds for each cluster.

In SLPA each nodes of the similarity graph maintains a memory of observed labels over time.
At iteration $t$, a node $i$ (speaker) selects a label from its
memory according to a speaking rule (typically the most frequent label)
and sends it to a randomly chosen neighbor $j$ (listener).
Formally, let $M_i^{(t)}$ denote the label memory of node $i$.
The speaker samples a label
$ \ell \sim P(\ell \mid M_i^{(t)})$
and transmits it to node $j$.
The listener then updates its memory as
$ M_j^{(t+1)} = M_j^{(t)} \cup \{\ell\}$.
After convergence, the probability of label $\ell$ at node $i$ is
$P_i(\ell) = \frac{\text{count}_i(\ell)}{|M_i|}$.
A threshold is applied to remove weak memberships on the above calculated probability.
This produces a set of overlapping communities
$ \mathcal{C} = \{C_1, C_2, \dots, C_m\}$
which naturally capture shared viewpoints across image groups.
To ensure connectivity between clusters, we enforce a minimum overlap
constraint. \\

\textbf{\textit{Cluster Graph Construction, Cluster-wise sparse SfM, and Hierarchical Cluster Map Merging: }} We build a cluster graph $G_c = (V_c, E_c)$ where nodes denote clusters and edge weights $w_{ij} = |C_i \cap C_j|$ capture shared images. A Minimum Spanning Tree (MinST) (via Kruskal) defines the merging order. The base cluster is selected by solving a Minimum Height Tree (MHT) on the MinST through iterative leaf removal, choosing the remaining node (or the larger one if two remain). Clusters are then merged hierarchically from leaves to the base.

Within each cluster, feature matching is performed using MASt3R~\cite{leroy2024groundingimagematching3d} restricted to edges in the original similarity graph, and sparse reconstruction is obtained using COLMAP~\cite{schoenberger2016sfm}. Inter-cluster alignment is achieved by estimating a $\text{Sim}(3)$ transformation (rotation, translation, scale) via RANSAC on shared camera poses. The source reconstruction is transformed into the target frame, followed by track merging based on shared observations and removal of high reprojection error outliers to ensure consistency.

\subsection{Sparse SFM to Mesh Surface}

\textbf{\textit{Area partitioning of large-scale sparse SfM: }}
\label{sec:area_decomp_ranking}
To scale reconstruction to large scenes, we first spatially partition the sparse COLMAP point cloud and then select, for each partition, a compact set of informative cameras. We parameterize the scene using a dominant support plane, assuming only a single dominant supporting surface rather than Manhattan-world structure. Projecting the sparse 3D points onto this plane yields a 2D footprint that is better suited to area-based decomposition than the original SfM frame. We then rotate the in-plane coordinates to compact the footprint, divide the resulting domain into a regular grid, and slightly enlarge each cell to define the effective partition support. This introduces moderate overlap between neighboring partitions and improves robustness near partition boundaries. Full details of support-plane fitting, planar parameterization, orientation refinement, and inflated grid construction are provided in the supplementary (Sec.~\ref{sec:supp_area_partitioning}).

For partition $(r,c)$ with sparse support $\mathcal{P}_{r,c}$, we define the candidate camera pool as $\mathcal{I}^{\mathrm{cand}}_{r,c}=\bigcup_{p\in\mathcal{P}_{r,c}}\mathcal{I}(p)$, namely, all cameras that observe at least one point in the partition. Rather than ranking cameras individually, we rank \emph{co-visible camera pairs}, since local reconstruction quality depends on both visibility and geometric complementarity.
We retain only camera pairs with sufficient local co-visibility within the partition and assign each admissible pair a prior that favors moderate relative viewpoint change and strong local overlap. For each partition point $p$, each admissible observing pair is further scored using two point-dependent cues: triangulation quality and image-plane centrality. Thus, preferred pairs are well supported within the partition, provide a useful baseline for $p$, and observe $p$ away from image boundaries. To avoid over-counting many near-redundant pairs around the same point, we retain only the top-$K$ pairs for each point and normalize their scores into a unit vote. Each selected pair then splits its vote equally between its two cameras, and votes are accumulated over all partition points to obtain a camera score $S_i$ for each candidate view. Finally, the top $M$ cameras form the partition-specific subset $\mathcal{I}^{\mathrm{top}}_{r,c}$.

Compared with simple visibility counting, this ranking favors cameras that are not only present in a partition, but also geometrically informative for local reconstruction. Detailed definitions, scoring functions, and normalization steps are provided in supplementary material Sec.~\ref{sec:supp_camera_ranking}. \\

\noindent
\textbf{\textit{Dense Reconstruction and Surface Initialization:-}}
\label{sec:dense_init}

For each partition, we retain the top-$M$ cameras from the ranking stage and fix their SfM intrinsics and poses. We then densify the partition using MASt3R depth predictions and pixel correspondences~\cite{leroy2024groundingimagematching3d}, reusing cached outputs from the original similarity graph whenever possible. As a result, only pairs newly introduced by the final partitioning, typically near former inter-cluster boundaries, require additional inference. To address residual inter-view scale mismatch and local geometric noise in the per-view dense predictions, we perform a two-stage geometry alignment procedure inspired by MASt3R-SfM \cite{duisterhof2024mast3rsfmfullyintegratedsolutionunconstrained}, but tailored to fixed SfM cameras. We first estimate one global scale per view to align matched dense 3D points across overlapping images, then refine the dense geometry by enforcing multi-view reprojection consistency under the frozen cameras. This removes the dominant scale inconsistency, suppresses local depth errors, and improves cross-view agreement. The full formulation is deferred to the supplementary (\Cref{sec:supp_dense_init}).

Finally, we fuse the aligned depth maps with TSDF integration, extract oriented points from the fused volume, and reconstruct a watertight mesh using Screened Poisson Surface Reconstruction. The resulting mesh provides the initialization for the subsequent differentiable-rendering-based refinement stage. Overall, the module transforms sparse SfM cameras and largely cached MASt3R predictions into a dense, globally aligned, and denoised surface initialization without re-estimating camera poses. \\

\noindent
\textbf{\textit{Mesh Refinement: }}
\label{sec:mesh_refinement} Starting from the watertight initial mesh \(M_0=(V_0,F_0)\), we refine the surface by directly optimizing the mesh using losses based on differentiable rendering. For view \(j\), let \(S_j(u)\in[0,1]\) be the target silhouette and \(\hat S_j(u;M^k)\) the rendered silhouette. The silhouette loss is
\begin{equation}
\mathcal L_{\mathrm{sil}}^{(j)}(M^k)=
\frac{1}{|\Omega_{\mathrm{img}}|}
\sum_{u\in\Omega_{\mathrm{img}}}
\big(\hat S_j(u;M^k)-S_j(u)\big)^2.
\end{equation}
Let \(\mathcal N_j(u)\in\mathbb S^2\) be the predicted unit normal map in camera coordinates from a pretrained off-the-shelf monocular normal estimator \cite{wang2025mogeunlockingaccuratemonocular} and \(\hat{\mathcal N}_j(u;M^k)\) the rendered unit normal map. On the foreground support \(\Omega_j=\{u\mid S_j(u)=1\}\), the normal-map loss is
\begin{equation}
\mathcal L_n^{(j)}(M^k)=
\frac{1}{|\Omega_j|}
\sum_{u\in\Omega_j}
\left(
1-\hat{\mathcal N}_j(u;M^k)^\top \mathcal N_j(u)
\right).
\end{equation}

At iteration \(k\), the current mesh is \(M^k=(V^k,F^k)\). Given calibrated cameras \(\{K_j,T_{wc,j}\}_{j=1}^{N}\), we optimize
\begin{equation}
\Phi(M^k)=
\sum_{j=1}^{N}
\left(
\lambda_n\,\mathcal L_n^{(j)}(M^k)+
\lambda_s\,\mathcal L_{\mathrm{sil}}^{(j)}(M^k)
\right)+
\mathcal R(M^k),
\end{equation}
where \(\mathcal R\) is a lightweight mesh regularizer, e.g.\ Laplacian smoothing.

Gradients of \(\Phi\) are backpropagated through the renderer to vertex positions, which are updated using the Isotropic Adam optimizer~\cite{continuousremeshing}. The optimizer also provides a relative vertex velocity \(\nu^k(i)\in[0,1]\)~\cite{continuousremeshing}, which measures the normalized update magnitude at vertex \(i\) and serves as a local indicator of stabilization.

Our mesh optimization method adapts Continuous-Remeshing~\cite{continuousremeshing}, originally designed for object-level remeshing, to urban scene mesh optimization. A fixed vertex budget tends to under-refine detailed regions, while allowing the mesh to grow freely during alternating optimization and remeshing can rapidly increase the number of vertices and faces, leading to unstable memory usage, occasional out-of-memory failures, and slower iterations. More importantly, uniform refinement spends vertices where they are not needed, whereas concentrating them in detail-rich regions improves approximation quality under the same budget and can also improve optimization stability by keeping the number of optimizable parameters smaller. To address this, instead of the edge-length consistency constraint in~\cite{continuousremeshing}, we use a curvature-guided terminal target and a speed-aware slack schedule, which are better suited to large, spatially heterogeneous scenes.

Our key idea is to decouple where fine resolution is needed from when that refinement is released. To determine \emph{where}, we estimate a screen-space normal-variation field from the predicted normal maps. For each view \(j\), let \(N^{(j)}(u,v)\in\mathbb R^{3\times 2}\) denote the image-space Jacobian of the unit normal field. We define the local normal-rotation rate as \(s^{(j)}(u,v)=\|N^{(j)}(u,v)\|_2\), and convert a normal-rotation tolerance \(\theta_0\) into a projected target edge length
\begin{equation}
p_{\mathrm{tgt}}^{(j)}(u,v)=
\mathrm{clip}\!\left(\frac{\theta_0}{s^{(j)}(u,v)},\,p_{\min},\,p_{\max}\right).
\label{eq:main_pixel_target}
\end{equation}
Thus, regions with rapid normal variation receive smaller projected target edges. For a vertex \(i\) visible in view \(j\), with projection \(\pi^{(j)}(\mathbf v_i)\), depth \(z_i^{(j)}\), and focal length \(f^{(j)}\), this yields a world-space target \(L_{ij}=\frac{z_i^{(j)}}{f^{(j)}}\,p_{\mathrm{tgt}}^{(j)}(\pi^{(j)}(\mathbf v_i))\). Pooling across visible views \(\mathcal V_i\) gives the per-vertex curvature-guided baseline \(L_{\mathrm{ref\text{-}curv}}(i)=\operatorname*{median}_{j\in\mathcal V_i} L_{ij}\).

To determine \emph{when} refinement is allowed, we introduce a nonnegative slack \(\zeta^k(i)\) on top of this baseline and update it using the relative vertex velocity:
\begin{equation}
\zeta^{k}(i)\leftarrow
\max\!\Big(0,\zeta^{k-1}(i)+g\big(\nu^k(i)-\nu_{\mathrm{ref}}\big)\Big).
\label{eq:main_slack_update}
\end{equation}
The final per-vertex reference edge length is
\begin{equation}
L_{\mathrm{ref}}^{k}(i)=
\mathrm{clip}\!\Big(L_{\mathrm{ref\text{-}curv}}^{k}(i)+\zeta^k(i),\,L_{\min},\,L_{\max}\Big).
\label{eq:main_ref_vertex}
\end{equation}
Edge targets are derived from the vertex field during remeshing; details, initialization, and implementation choices are given in \cref{subsec:supp_curvature_field,subsec:supp_slack,subsec:supp_bounds,subsec:supp_remeshing}. Intuitively, curvature specifies the terminal resolution, while velocity delays refinement until local motion has sufficiently stabilized.

After each vertex update, we invoke continuous remeshing under the current reference field \(L_{\mathrm{ref}}^k\): short edges are collapsed, long edges are split, and flips improve local triangle quality and valence regularity, using the same validity checks as in~\cite{continuousremeshing}; the exact rules are given in \cref{subsec:supp_remeshing}. The global bounds \(L_{\min}\) and \(L_{\max}\) are estimated automatically from camera intrinsics, robust depth statistics, and current mesh scale as described in \cref{subsec:supp_bounds}.

The refinement therefore alternates render \(\rightarrow\) compare \(\rightarrow\) update \(\rightarrow\) remesh, and the final mesh is selected as the best snapshot \(M^\star=\arg\min_k \Phi(M^k)\).

Our adaptation of~\cite{continuousremeshing} using a curvature-guided terminal target and a speed-aware slack schedule concentrates triangles in high-curvature regions while keeping flatter regions coarse, making remeshing more stable and memory-efficient for large heterogeneous urban scenes. \\

\noindent
\textbf{\textit{Mesh Stitching: }} To integrate two overlapping surface meshes $M_1$ and $M_2$, we use a decomposition-reconstruction strategy that preserves high-frequency geometric detail while ensuring a watertight transition in the overlap region. For the meshes $M_1$ and $M_2$ with bounding volumes $B_1$ and $B_2$ obtained from the preceding partitioning stage (Sec.~\ref{sec:area_decomp_ranking}), the decomposition step extracts the non-overlapping exterior regions by clipping each mesh against the other’s volume:
\begin{equation}
    M_{1 \setminus 2} = M_1 \setminus B_2, \quad M_{2 \setminus 1} = M_2 \setminus B_1
\end{equation}
where $\setminus$ denotes clipping against the opposing bounding volume.
To resolve the discontinuity at the intersection, we extract the overlap region $\mathcal{O} = B_1 \cap B_2$. Let $P_{int}$ be the set of vertices residing in the overlap region. We reconstruct the bridging surface $M_{seam}$ by applying a Delaunay triangulation $D$ over the projected manifold, i.e., $M_{seam} = D(P_{int})$.

The final merged mesh $M_{final}$ is obtained by combining the preserved exterior surfaces with the reconstructed seam patch, followed by a stitching operation along their boundary vertices:
\begin{equation}
    M_{final} = \mathrm{Stitch}(M_{1 \setminus 2},\, M_{2 \setminus 1},\, M_{seam})
\end{equation}
where $\mathrm{Stitch}(\cdot)$ denotes a boundary-aware topological refinement operator that aligns and welds corresponding boundary vertices within a distance threshold $\epsilon$, removes duplicate or degenerate faces, and enforces manifold connectivity across the three mesh components. Please refer to \cref{fig:MeshMerge} in supplementary, for mesh stitching visualization.

\section{Experiments and Results}

\subsection{Datasets and Implementation Details}
Our experiments cover five representative scenes drawn from two datasets: CUHK-LOWER, CUHK-UPPER, LFLS, SZIIT from GauU-Scene \cite{xiong2024gauuscenescenereconstructionbenchmark} and Residence from UrbanScene3D \cite{lin2022capturingreconstructingsimulatingurbanscene3d}. 

Our method was implemented on a (2 x 48 GB) NVIDIA A6000 Ada GPU system and all methods including ours were evaluated on the same system.

\begin{table}[t]
\centering
\scriptsize
\setlength{\tabcolsep}{4pt}
\begin{tabular}{llcccc}
\toprule
\textbf{Scene} & \textbf{Method} & \textbf{Precision $\uparrow$} & \textbf{Recall $\uparrow$} & \textbf{F1 $\uparrow$} & \textbf{Time (min) $\downarrow$} \\
\midrule

\multirow{3}{*}{CUHK-LOWER}
& CityGS-v2 & 0.1312 & \textcolor{blue}{0.0820} & \textcolor{blue}{0.1009} & 340.90 \\
& CityGS-X  & \textcolor{blue}{0.1327} & 0.0758 & 0.0965 & \textcolor{ForestGreen}{\textbf{75.00}} \\
& Ours      & \textcolor{ForestGreen}{\textbf{0.1420}} & \textcolor{ForestGreen}{\textbf{0.0910}} & \textcolor{ForestGreen}{\textbf{0.1110}} & \textcolor{blue}{95.00} \\
\midrule

\multirow{3}{*}{CUHK-UPPER}
& CityGS-v2 & 0.2203 & 0.0953 & 0.1330 & 371.05 \\
& CityGS-X  & \textcolor{ForestGreen}{\textbf{0.2950}} & \textcolor{ForestGreen}{\textbf{0.1510}} & \textcolor{ForestGreen}{\textbf{0.2000}} & \textcolor{ForestGreen}{\textbf{74.00}} \\
& Ours      & \textcolor{blue}{0.2680} & \textcolor{blue}{0.1120} & \textcolor{blue}{0.1580} & \textcolor{blue}{113.00} \\
\midrule

\multirow{3}{*}{LFLS}
& CityGS-v2 & 0.0901 & \textcolor{blue}{0.0666} & \textcolor{blue}{0.0791} & 296.90 \\
& CityGS-X  & \textcolor{ForestGreen}{\textbf{0.0978}} & 0.0599 & 0.0743 & \textcolor{blue}{87.00} \\
& Ours      & \textcolor{blue}{0.0951} & \textcolor{ForestGreen}{\textbf{0.0777}} & \textcolor{ForestGreen}{\textbf{0.0855}} & \textcolor{ForestGreen}{\textbf{83.00}} \\
\midrule

\multirow{3}{*}{SZIIT}
& CityGS-v2 & 0.1513 & 0.0488 & 0.0738 & 299.07 \\
& CityGS-X  & \textcolor{blue}{0.1849} & \textcolor{blue}{0.0538} & \textcolor{blue}{0.0833} & \textcolor{blue}{94.00} \\
& Ours      & \textcolor{ForestGreen}{\textbf{0.1925}} & \textcolor{ForestGreen}{\textbf{0.0647}} & \textcolor{ForestGreen}{\textbf{0.0968}} & \textcolor{ForestGreen}{\textbf{92.00}} \\
\bottomrule

\end{tabular}
\caption{Quantitative comparison of surface reconstruction performance on the GauU-Scene dataset against recent city-scale methods \cite{gao2025citygs,liu2024citygaussianv2}. Best results are highlighted in \textcolor{ForestGreen}{\textbf{green bold}}, second-best in \textcolor{blue}{blue}}
\label{tab:sota_comparison}
\end{table}

\begin{table}[t]
\centering
\footnotesize
\setlength{\tabcolsep}{2pt}
\begin{tabular}{l c c}
\hline
\textbf{Method} & \textbf{Time (hrs)$\downarrow$} & \textbf{2D Rep. Err. (px) $\downarrow$} \\
\hline
MASt3R-COLMAP (all images)         & 53.88                                  & \textcolor{ForestGreen}{\textbf{1.17}} \\
MASt3R-GLOMAP (all images)         & \textcolor{blue}{19.45}                & 1.856                                  \\
Ours (Clustering + MASt3R-COLMAP)  & \textcolor{ForestGreen}{\textbf{2.74}} & \textcolor{blue}{1.38}                 \\
\hline
\end{tabular}
\caption{Comparison of our clustering-based SFM strategy with state-of-the-art SFM methods on the GauU-Scene dataset. MASt3R-COLMAP and MASt3R-GLOMAP denote pipelines where MASt3R is used for feature matching frontend, and COLMAP or GLOMAP are employed as the SfM mapper backend for sparse reconstruction.
}
\label{tab:colmap_glomap}
\end{table}

\begin{figure*}[ht!] 
    \centering
    \includegraphics[width=\textwidth]{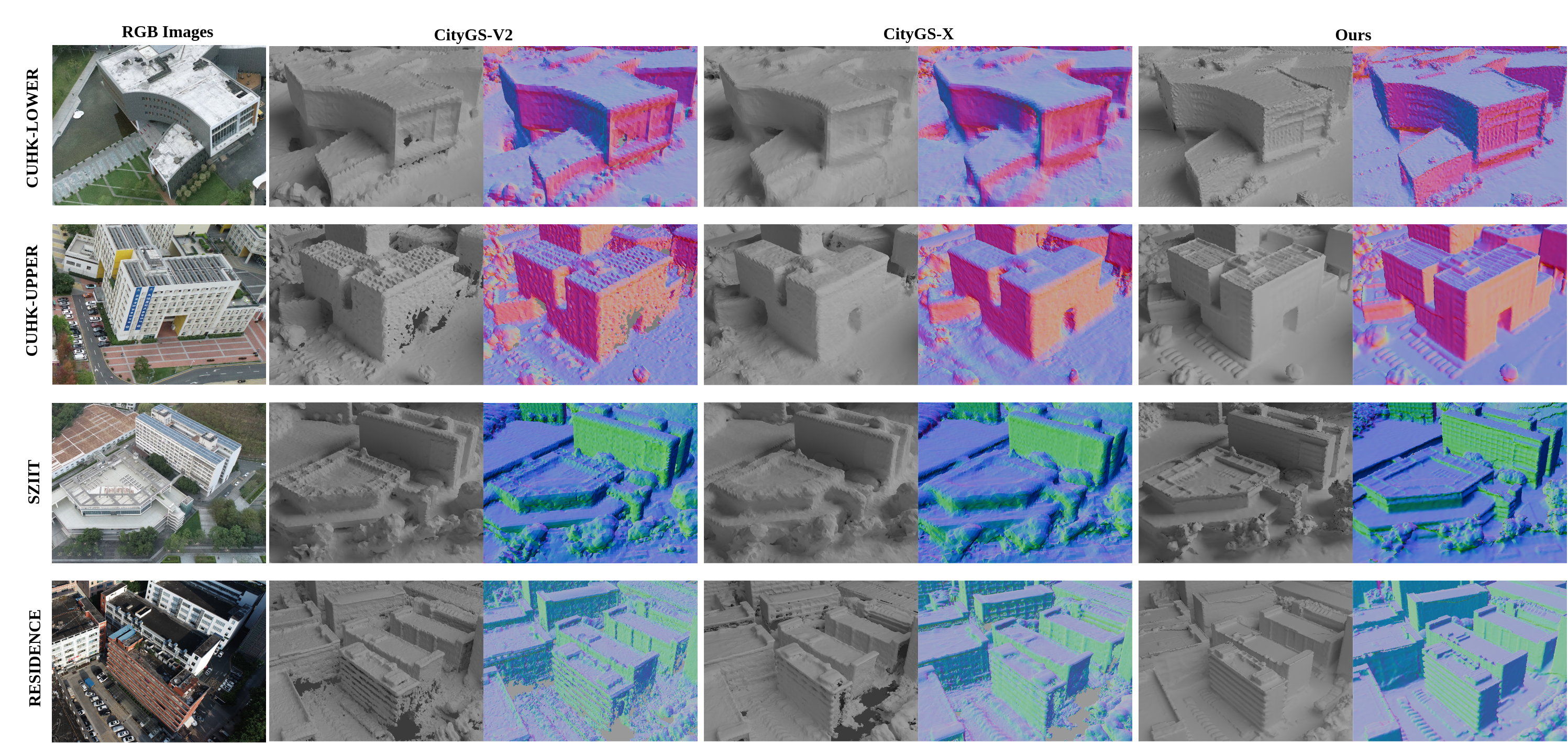}
    \caption{Qualitative Comparison with recent city-scale surface reconstruction methods (CityGS-v2\cite{liu2024citygaussianv2} and CityGS-X\cite{gao2025citygs}) on GauU-Scene\cite{xiong2024gauu} dataset (CUHK-LOWER, CUHK-UPPER, SZIIT) and UrbanScene3D\cite{lin2022capturingreconstructingsimulatingurbanscene3d} dataset (Residence). Left one is the reference RGB image, the mesh and normals were rendered from nearby novel views. For more results, please refer to supplementary \cref{fig:Qual_supply} }
    \label{fig:Qual_main}
\end{figure*}

\subsection{Quantitative Comparison}

As shown in Table~\ref{tab:sota_comparison}, our full method achieves a strong trade-off between reconstruction quality and runtime across diverse large-scale scenes. It attains the best overall performance on CUHK-LOWER and SZIIT, where it achieves the highest Precision, Recall, and F1-score, and also performs best on LFLS in terms of Recall, F1-score, and runtime. Relative to CityGS-v2 \cite{liu2024citygaussianv2}, our method consistently improves both accuracy and efficiency, delivering higher geometric quality with substantially lower runtime on all evaluated scenes. Relative to CityGS-X \cite{gao2025citygs}, our method produces better reconstruction quality on most scenes while retaining comparable execution time. Although CityGS-X performs best on CUHK-UPPER, our approach remains competitive and continues to offer a significantly better runtime than CityGS-v2. Overall, these results demonstrate that our pipeline is an effective and scalable solution for city-scale surface reconstruction. 
In Table~\ref{tab:colmap_glomap}, we also demonstrate the effectiveness of our proposed \textit{Image to Sparse SfM} stage. We achieve a significantly lower reprojection error compared to MASt3R-GLOMAP, while being substantially more efficient than MASt3R-COLMAP, which requires hours despite achieving a slightly lower error. This highlights that our approach provides a favorable trade-off between accuracy and computational cost, bridging the gap between highly accurate but slow incremental reconstruction and fast but less accurate global methods.

\subsection{Qualitative Comparison}

\begin{figure*}[ht!] 
    \centering
    \includegraphics[width=\textwidth]{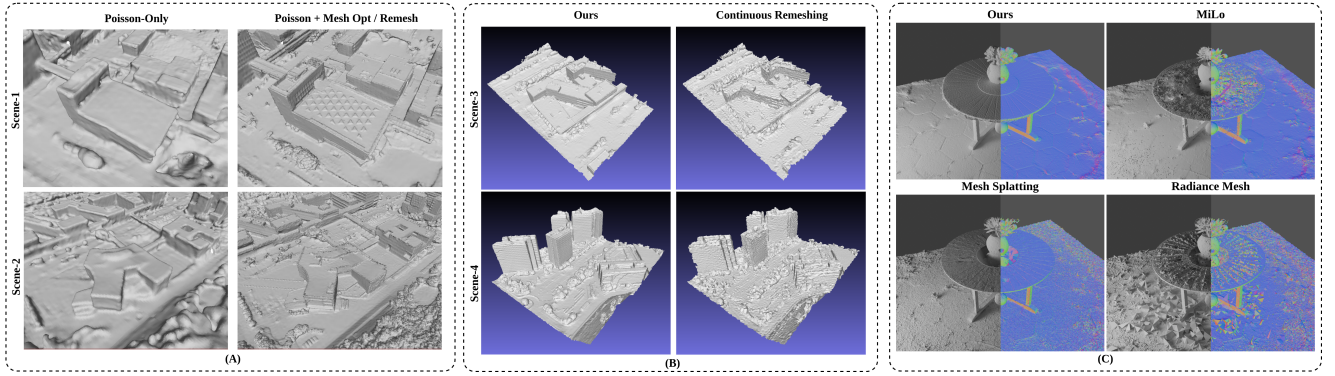}
    \caption{(A) Ablation Study of "Poisson Only" vs "Poisson + MeshOpt/Remesh" across different scenes of GauU-Scene dataset \\ (B) Qualitative Comparison with Continuous-Remeshing \cite{continuousremeshing} across different scenes of GauU-Scene dataset\cite{xiong2024gauu} under fixed maximum vertex budget. (C) Qualitative comparison of surface reconstruction methods on the Garden scene from the MipNeRF-360 \cite{barron2022mipnerf360} dataset. As evident from the mesh/normal renders, our novel surface reconstruction method produces cleaner geometry, sharper structural boundaries, and more watertight surfaces compared to MiLo \cite{guedon2025milo}, Mesh Splatting \cite{Held2025MeshSplatting}, and Radiance Mesh \cite{mai2025radiancemeshesvolumetricreconstruction}. For mesh quality metric comparison, refer to supplementary \cref{tab:mesh_quality}.}
    \label{multi_table_poisson_conremesh_milo}
\end{figure*}

Qualitative comparisons in Fig.~\ref{fig:Qual_main} further show that our method produces a watertight mesh surface with lower artifacts, smooth surface normals, and detailed geometry when compared to the latest methods of large-scale 3D reconstruction. We further evaluate multiple small-scale surface reconstruction methods \cite{guedon2025milo, Held2025MeshSplatting, mai2025radiancemeshesvolumetricreconstruction} with our method. As shown in  ~\cref{multi_table_poisson_conremesh_milo} (C), qualitative results demonstrate that our approach outperforms existing small-scale methods.

\subsection{Ablation Study}
\label{sec:surf_ablation_study}

\textbf{Our Proposed Image Clustering Method vs Existing Methods}
Ablation of different clustering methods in Table~\ref{tab:clustering_methods} shows that our clustering method (SLPA) consistently outperforms both N-cut \cite{chen2020graphbasedparallellargescale} and agglomerative clustering \cite{scikit-learn} across different scenes. Specifically, our method achieves the lowest 2D reprojection error and significantly reduces the percentage of image rejections during SfM compared to competing approaches. 
These results demonstrate that our clustering strategy produces more coherent and reliable image groupings, which directly contribute to improved reconstruction quality and stability.\\

\textbf{Poisson Surface Reconstruction vs.\ Our Mesh Refinement.}
To evaluate the importance of the mesh refinement, we compare the mesh obtained directly from Screened Poisson reconstruction (before mesh refinement) with the output of our full pipeline. As reported in Table~\ref{tab:poisson_ablation}, our proposed mesh refinement consistently improves Precision, Recall, and F1 on both scenes. 
\cref{multi_table_poisson_conremesh_milo} (A) demonstrates the clear improvement in surface details captured by the proposed mesh refinement stage when compared to the output from Poisson mesh reconstruction.\\

\begin{table}[t]
\centering
\footnotesize
\setlength{\tabcolsep}{2pt}
\resizebox{\columnwidth}{!}{
\begin{tabular}{l c c}
\hline
\textbf{Method} & \textbf{2D Reproj. Err. $\downarrow$} & \textbf{\% Img. Rej. $\downarrow$} \\
\hline

Our Clustering (SLPA)     & \textcolor{ForestGreen}{\textbf{1.38}} & \textcolor{ForestGreen}{\textbf{4.373}} \\
N-cut Clustering          & \textcolor{blue}{1.441}                & \textcolor{blue}{9.78}                  \\
Agglomerative Clustering  & 1.563                                  & 10.27                                   \\
\hline
\end{tabular}
}
\caption{Comparison of clustering methods after cluster-wise SfM and map merging on the GauU-Scene dataset.}
\label{tab:clustering_methods}
\end{table}

\begin{table}[t]
\centering
\footnotesize
\setlength{\tabcolsep}{2pt}
\begin{tabular}{l l c c c}
\hline
\textbf{Scene} & \textbf{Method} & \textbf{P $\uparrow$} & \textbf{R $\uparrow$} & \textbf{F1 $\uparrow$} \\
\hline

\multirow{2}{*}{CUHK-LOWER}
& Ours (Poisson Only)         & 0.131                                  & 0.049                                  & 0.071 \\
& Ours (Poisson + Opt/Remesh) & \textcolor{ForestGreen}{\textbf{0.142}} & \textcolor{ForestGreen}{\textbf{0.091}} & \textcolor{ForestGreen}{\textbf{0.111}} \\
\hline

\multirow{2}{*}{SZIIT}
& Ours (Poisson Only)         & 0.1321                                 & 0.0163                                 & 0.029 \\
& Ours (Poisson + Opt/Remesh) & \textcolor{ForestGreen}{\textbf{0.1925}} & \textcolor{ForestGreen}{\textbf{0.0647}} & \textcolor{ForestGreen}{\textbf{0.0968}} \\
\hline

\end{tabular}
\caption{Ablation of Poisson only vs.\ Poisson + Opt/Remesh on selected GauU-Scene scenes. }
\label{tab:poisson_ablation}
\end{table}

\begin{table}[t]
\centering
\footnotesize
\setlength{\tabcolsep}{3pt}
\begin{tabular}{l l c c c}
\hline
\textbf{Scene-Name} & \textbf{Method} & \textbf{Precision $\uparrow$} & \textbf{Recall $\uparrow$} & \textbf{F1 $\uparrow$} \\
\hline

\multirow{2}{*}{CUHK-LOWER}
& Ours      & \textcolor{ForestGreen}{\textbf{0.142}}  & \textcolor{ForestGreen}{\textbf{0.091}}  & \textcolor{ForestGreen}{\textbf{0.111}}  \\
& ConRemesh & 0.1350                                    & 0.0632                                    & 0.0861                                    \\
\hline

\multirow{2}{*}{SZIIT}
& Ours      & \textcolor{ForestGreen}{\textbf{0.1925}} & \textcolor{ForestGreen}{\textbf{0.0647}} & \textcolor{ForestGreen}{\textbf{0.0968}} \\
& ConRemesh & 0.1858                                    & 0.0317                                    & 0.0542                                    \\
\hline
\end{tabular}
\caption{Quantitative comparison of Ours vs.\ original Continuous-Remeshing across different scenes.}
\label{tab:conremesh_comparison}
\end{table}

\textbf{Continuous Remeshing vs.\ Our Method.}
We further compare our mesh refinement algorithm with the original formulation of Continuous Remeshing \cite{continuousremeshing} under a fixed maximum vertex-budget cap, so that both methods operate under the same mesh-complexity constraint.~\cref{tab:conremesh_comparison} and \cref{multi_table_poisson_conremesh_milo} (B) shows that our method consistently achieves better results across both scenes.
This is because near-uniform remeshing spends vertices unnecessarily in flat regions, leaving less budget for high-curvature areas and fine structures. In contrast, our method adaptively redistributes the same vertex budget using curvature-guided edge-length targets and a speed-aware refinement schedule, enabling more accurate reconstruction where detail is most needed.

\section{Conclusion}

We presented City-Mesh3R, an end-to-end scalable pipeline reconstructs simulation-ready 3D meshes from unordered images of large urban scenes. City-Mesh3R combines a distributed hierarchical sparse-to-dense design, a two-stage partitioning strategy for sparse SfM and dense reconstruction, and a curvature-aware adaptive remeshing module for high-fidelity surface recovery while maintaining topological consistency across independently processed clusters. Experimental results on urban-scale 3D datasets demonstrate a significant performance gain achieved by our method both in processing time, and improved 3D surface quality, important for applications in urban planning, infrastructure monitoring, etc.

{
    \small
    \bibliographystyle{ieeenat_fullname}
    \bibliography{main}

@String(CVPR= {IEEE Conf. Comput. Vis. Pattern Recog.})

@String(ICCV= {Int. Conf. Comput. Vis.})

@String(ECCV= {Eur. Conf. Comput. Vis.})

@String(TOG= {ACM Trans. Graph.})

@String(AAAI = {AAAI})

@String(CVPR  = {CVPR})

@String(ICCV  = {ICCV})

@String(ECCV  = {ECCV})

@String(TOG   = {ACM TOG})

@misc{lin2022capturingreconstructingsimulatingurbanscene3d,
      title={Capturing, Reconstructing, and Simulating: the UrbanScene3D Dataset}, 
      author={Liqiang Lin and Yilin Liu and Yue Hu and Xingguang Yan and Ke Xie and Hui Huang},
      year={2022},
      eprint={2107.04286},
      archivePrefix={arXiv},
      primaryClass={cs.CV},
      url={https://arxiv.org/abs/2107.04286}, 
}

@misc{xiong2024gauuscenescenereconstructionbenchmark,
      title={GauU-Scene: A Scene Reconstruction Benchmark on Large Scale 3D Reconstruction Dataset Using Gaussian Splatting}, 
      author={Butian Xiong and Zhuo Li and Zhen Li},
      year={2024},
      eprint={2401.14032},
      archivePrefix={arXiv},
      primaryClass={cs.CV},
      url={https://arxiv.org/abs/2401.14032}, 
}

@misc{leroy2024groundingimagematching3d,
      title={Grounding Image Matching in 3D with MASt3R}, 
      author={Vincent Leroy and Yohann Cabon and Jérôme Revaud},
      year={2024},
      eprint={2406.09756},
      archivePrefix={arXiv},
      primaryClass={cs.CV},
      url={https://arxiv.org/abs/2406.09756}, 
}

@misc{duisterhof2024mast3rsfmfullyintegratedsolutionunconstrained,
      title={MASt3R-SfM: a Fully-Integrated Solution for Unconstrained Structure-from-Motion}, 
      author={Bardienus Duisterhof and Lojze Zust and Philippe Weinzaepfel and Vincent Leroy and Yohann Cabon and Jerome Revaud},
      year={2024},
      eprint={2409.19152},
      archivePrefix={arXiv},
      primaryClass={cs.CV},
      url={https://arxiv.org/abs/2409.19152}, 
}

@article{scikit-learn,
  title={Scikit-learn: Machine Learning in {P}ython},
  author={Pedregosa, F. and Varoquaux, G. and Gramfort, A. and Michel, V.
          and Thirion, B. and Grisel, O. and Blondel, M. and Prettenhofer, P.
          and Weiss, R. and Dubourg, V. and Vanderplas, J. and Passos, A.
          and Cournapeau, D. and Brucher, M. and Perrot, M. and Duchesnay, E.},
  journal={Journal of Machine Learning Research},
  volume={12},
  pages={2825--2830},
  year={2011}
}

@article{xie2011slpa,
  title={SLPA: Uncovering overlapping communities in social networks via a speaker-listener interaction dynamic process},
  author={Xie, Jierui and Szymanski, Boleslaw K},
  journal={arXiv preprint arXiv:1109.5720},
  year={2011}
}

@misc{oquab2024dinov2learningrobustvisual,
      title={DINOv2: Learning Robust Visual Features without Supervision}, 
      author={Maxime Oquab and Timothée Darcet and Théo Moutakanni and Huy Vo and Marc Szafraniec and Vasil Khalidov and Pierre Fernandez and Daniel Haziza and Francisco Massa and Alaaeldin El-Nouby and Mahmoud Assran and Nicolas Ballas and Wojciech Galuba and Russell Howes and Po-Yao Huang and Shang-Wen Li and Ishan Misra and Michael Rabbat and Vasu Sharma and Gabriel Synnaeve and Hu Xu and Hervé Jegou and Julien Mairal and Patrick Labatut and Armand Joulin and Piotr Bojanowski},
      year={2024},
      eprint={2304.07193},
      archivePrefix={arXiv},
      primaryClass={cs.CV},
      url={https://arxiv.org/abs/2304.07193}, 
}

@misc{revaud2019learningaverageprecisiontraining,
      title={Learning with Average Precision: Training Image Retrieval with a Listwise Loss}, 
      author={Jerome Revaud and Jon Almazan and Rafael Sampaio de Rezende and Cesar Roberto de Souza},
      year={2019},
      eprint={1906.07589},
      archivePrefix={arXiv},
      primaryClass={cs.CV},
      url={https://arxiv.org/abs/1906.07589}, 
}

@misc{chen2020graphbasedparallellargescale,
      title={Graph-Based Parallel Large Scale Structure from Motion}, 
      author={Yu Chen and Shuhan Shen and Yisong Chen and Guoping Wang},
      year={2020},
      eprint={1912.10659},
      archivePrefix={arXiv},
      primaryClass={cs.CV},
      url={https://arxiv.org/abs/1912.10659}, 
}

@misc{sarlin2019coarsefinerobusthierarchical,
      title={From Coarse to Fine: Robust Hierarchical Localization at Large Scale}, 
      author={Paul-Edouard Sarlin and Cesar Cadena and Roland Siegwart and Marcin Dymczyk},
      year={2019},
      eprint={1812.03506},
      archivePrefix={arXiv},
      primaryClass={cs.CV},
      url={https://arxiv.org/abs/1812.03506}, 
}

@misc{humenberger2022robustimageretrievalbasedvisual,
      title={Robust Image Retrieval-based Visual Localization using Kapture}, 
      author={Martin Humenberger and Yohann Cabon and Nicolas Guerin and Julien Morat and Vincent Leroy and Jérôme Revaud and Philippe Rerole and Noé Pion and Cesar de Souza and Gabriela Csurka},
      year={2022},
      eprint={2007.13867},
      archivePrefix={arXiv},
      primaryClass={cs.CV},
      url={https://arxiv.org/abs/2007.13867}, 
}

@misc{arandjelovic2016netvladcnnarchitectureweakly,
      title={NetVLAD: CNN architecture for weakly supervised place recognition}, 
      author={Relja Arandjelović and Petr Gronat and Akihiko Torii and Tomas Pajdla and Josef Sivic},
      year={2016},
      eprint={1511.07247},
      archivePrefix={arXiv},
      primaryClass={cs.CV},
      url={https://arxiv.org/abs/1511.07247}, 
}

@article{liu2024citygaussianv2,
  title={Citygaussianv2: Efficient and geometrically accurate reconstruction for large-scale scenes},
  author={Liu, Yang and Luo, Chuanchen and Mao, Zhongkai and Peng, Junran and Zhang, Zhaoxiang},
  journal={arXiv preprint arXiv:2411.00771},
  year={2024}
}

@inproceedings{liu2024citygaussian,
  title={Citygaussian: Real-time high-quality large-scale scene rendering with gaussians},
  author={Liu, Yang and Luo, Chuanchen and Fan, Lue and Wang, Naiyan and Peng, Junran and Zhang, Zhaoxiang},
  booktitle={European Conference on Computer Vision},
  pages={265--282},
  year={2024},
  organization={Springer}
}

@inproceedings{turki2022mega,
  title={Mega-nerf: Scalable construction of large-scale nerfs for virtual fly-throughs},
  author={Turki, Haithem and Ramanan, Deva and Satyanarayanan, Mahadev},
  booktitle={Proceedings of the IEEE/CVF conference on computer vision and pattern recognition},
  pages={12922--12931},
  year={2022}
}

@article{zhang2024supernerf,
  title={SuperNeRF: High-precision 3D reconstruction for large-scale scenes},
  author={Zhang, Guangyun and Xue, Chaozhong and Zhang, Rongting},
  journal={IEEE Transactions on Geoscience and Remote Sensing},
  year={2024},
  publisher={IEEE}
}

@inproceedings{lin2024vastgaussian,
  title={Vastgaussian: Vast 3d gaussians for large scene reconstruction},
  author={Lin, Jiaqi and Li, Zhihao and Tang, Xiao and Liu, Jianzhuang and Liu, Shiyong and Liu, Jiayue and Lu, Yangdi and Wu, Xiaofei and Xu, Songcen and Yan, Youliang and others},
  booktitle={Proceedings of the IEEE/CVF Conference on Computer Vision and Pattern Recognition},
  pages={5166--5175},
  year={2024}
}

@inproceedings{chen2025gigags,
  title={GigaGS: 3D Gaussian Based Planar Representation for Large-Scene Surface Reconstruction},
  author={Chen, Junyi and Ye, Weicai and Wang, Yifan and Chen, Danpeng and Huang, Di and Ouyang, Wanli and Zhang, Guofeng and Qiao, Yu and He, Tong},
  booktitle={Proceedings of the AAAI Conference on Artificial Intelligence},
  volume={39},
  number={2},
  pages={2088--2096},
  year={2025}
}

@inproceedings{jiang2025horizon,
  title={Horizon-GS: Unified 3D Gaussian Splatting for Large-Scale Aerial-to-Ground Scenes},
  author={Jiang, Lihan and Ren, Kerui and Yu, Mulin and Xu, Linning and Dong, Junting and Lu, Tao and Zhao, Feng and Lin, Dahua and Dai, Bo},
  booktitle={Proceedings of the Computer Vision and Pattern Recognition Conference},
  pages={26789--26799},
  year={2025}
}

@article{xiong2024sa,
  title={Sa-gs: Semantic-aware gaussian splatting for large scene reconstruction with geometry constrain},
  author={Xiong, Butian and Ye, Xiaoyu and Tse, Tze Ho Elden and Han, Kai and Cui, Shuguang and Li, Zhen},
  journal={arXiv preprint arXiv:2405.16923},
  year={2024}
}

@article{chen2024pgsr,
  title={Pgsr: Planar-based gaussian splatting for efficient and high-fidelity surface reconstruction},
  author={Chen, Danpeng and Li, Hai and Ye, Weicai and Wang, Yifan and Xie, Weijian and Zhai, Shangjin and Wang, Nan and Liu, Haomin and Bao, Hujun and Zhang, Guofeng},
  journal={IEEE Transactions on Visualization and Computer Graphics},
  year={2024},
  publisher={IEEE}
}

@article{xiong2024gauu,
  title={Gauu-scene: A scene reconstruction benchmark on large scale 3d reconstruction dataset using gaussian splatting},
  author={Xiong, Butian and Li, Zhuo and Li, Zhen},
  journal={arXiv preprint arXiv:2401.14032},
  year={2024}
}

@article{chen2024dogs,
  title={Dogs: Distributed-oriented gaussian splatting for large-scale 3d reconstruction via gaussian consensus},
  author={Chen, Yu and Lee, Gim Hee},
  journal={Advances in Neural Information Processing Systems},
  volume={37},
  pages={34487--34512},
  year={2024}
}

@article{gao2025citygs,
  title={Citygs-x: A scalable architecture for efficient and geometrically accurate large-scale scene reconstruction},
  author={Gao, Yuanyuan and Li, Hao and Chen, Jiaqi and Zou, Zhengyu and Zhong, Zhihang and Zhang, Dingwen and Sun, Xiao and Han, Junwei},
  journal={arXiv preprint arXiv:2503.23044},
  year={2025}
}

@inproceedings{tancik2022block,
  title={Block-nerf: Scalable large scene neural view synthesis},
  author={Tancik, Matthew and Casser, Vincent and Yan, Xinchen and Pradhan, Sabeek and Mildenhall, Ben and Srinivasan, Pratul P and Barron, Jonathan T and Kretzschmar, Henrik},
  booktitle={Proceedings of the IEEE/CVF conference on computer vision and pattern recognition},
  pages={8248--8258},
  year={2022}
}

@article{agarwal2011building,
  title={Building rome in a day},
  author={Agarwal, Sameer and Furukawa, Yasutaka and Snavely, Noah and Simon, Ian and Curless, Brian and Seitz, Steven M and Szeliski, Richard},
  journal={Communications of the ACM},
  volume={54},
  number={10},
  pages={105--112},
  year={2011},
  publisher={ACM New York, NY, USA}
}

@article{yang2025scalable,
  title={Scalable and high-quality neural implicit representation for 3D reconstruction},
  author={Yang, Leyuan and Deng, Bailin and Zhang, Juyong},
  journal={IEEE Transactions on Visualization and Computer Graphics},
  year={2025},
  publisher={IEEE}
}

@article{kerbl2024hierarchical,
  title={A hierarchical 3d gaussian representation for real-time rendering of very large datasets},
  author={Kerbl, Bernhard and Meuleman, Andreas and Kopanas, Georgios and Wimmer, Michael and Lanvin, Alexandre and Drettakis, George},
  journal={ACM Transactions on Graphics (TOG)},
  volume={43},
  number={4},
  pages={1--15},
  year={2024},
  publisher={ACM New York, NY, USA}
}

@article{li2025geosvr,
  title={GeoSVR: Taming Sparse Voxels for Geometrically Accurate Surface Reconstruction},
  author={Li, Jiahe and Zhang, Jiawei and Zhang, Youmin and Bai, Xiao and Zheng, Jin and Yu, Xiaohan and Gu, Lin},
  journal={arXiv preprint arXiv:2509.18090},
  year={2025}
}

@inproceedings{guedon2024sugar,
  title={Sugar: Surface-aligned gaussian splatting for efficient 3d mesh reconstruction and high-quality mesh rendering},
  author={Gu{\'e}don, Antoine and Lepetit, Vincent},
  booktitle={Proceedings of the IEEE/CVF Conference on Computer Vision and Pattern Recognition},
  pages={5354--5363},
  year={2024}
}

@inproceedings{huang20242d,
  title={2d gaussian splatting for geometrically accurate radiance fields},
  author={Huang, Binbin and Yu, Zehao and Chen, Anpei and Geiger, Andreas and Gao, Shenghua},
  booktitle={ACM SIGGRAPH 2024 conference papers},
  pages={1--11},
  year={2024}
}

@article{yu2024gaussian,
  title={Gaussian opacity fields: Efficient adaptive surface reconstruction in unbounded scenes},
  author={Yu, Zehao and Sattler, Torsten and Geiger, Andreas},
  journal={ACM Transactions on Graphics (ToG)},
  volume={43},
  number={6},
  pages={1--13},
  year={2024},
  publisher={ACM New York, NY, USA}
}

@inproceedings{sun2025sparse,
  title={Sparse Voxels Rasterization: Real-time High-fidelity Radiance Field Rendering},
  author={Sun, Cheng and Choe, Jaesung and Loop, Charles and Ma, Wei-Chiu and Wang, Yu-Chiang Frank},
  booktitle={Proceedings of the Computer Vision and Pattern Recognition Conference},
  pages={16187--16196},
  year={2025}
}

@inproceedings{gao2025surfacesplat,
  title={SurfaceSplat: Connecting Surface Reconstruction and Gaussian Splatting},
  author={Gao, Zihui and Bian, Jia-Wang and Lin, Guosheng and Chen, Hao and Shen, Chunhua},
  booktitle={Proceedings of the IEEE/CVF International Conference on Computer Vision},
  pages={28525--28534},
  year={2025}
}

@article{lyu20243dgsr,
  title={3dgsr: Implicit surface reconstruction with 3d gaussian splatting},
  author={Lyu, Xiaoyang and Sun, Yang-Tian and Huang, Yi-Hua and Wu, Xiuzhe and Yang, Ziyi and Chen, Yilun and Pang, Jiangmiao and Qi, Xiaojuan},
  journal={ACM Transactions on Graphics (TOG)},
  volume={43},
  number={6},
  pages={1--12},
  year={2024},
  publisher={ACM New York, NY, USA}
}

@inproceedings{zhu2018very,
  title={Very large-scale global sfm by distributed motion averaging},
  author={Zhu, Siyu and Zhang, Runze and Zhou, Lei and Shen, Tianwei and Fang, Tian and Tan, Ping and Quan, Long},
  booktitle={Proceedings of the IEEE conference on computer vision and pattern recognition},
  pages={4568--4577},
  year={2018}
}

@article{continuousremeshing,
author = {Palfinger, Werner},
title = {Continuous remeshing for inverse rendering},
journal = {Computer Animation and Virtual Worlds},
volume = {33},
number = {5},
pages = {e2101},
doi = {https://doi.org/10.1002/cav.2101},
url = {https://onlinelibrary.wiley.com/doi/abs/10.1002/cav.2101},
eprint = {https://onlinelibrary.wiley.com/doi/pdf/10.1002/cav.2101},
year = {2022}
}

@ARTICLE{CIty10892091,
  author={Christodoulides, Andreas and Tam, Gary K. L. and Clarke, James and Smith, Richard and Horgan, Jon and Micallef, Nicholas and Morley, Jeremy and Villamizar, Nelly and Walton, Sean},
  journal={IEEE Transactions on Visualization and Computer Graphics}, 
  title={Survey on 3D Reconstruction Techniques: Large-Scale Urban City Reconstruction and Requirements}, 
  year={2025},
  volume={31},
  number={10},
  pages={9343-9367},
  doi={10.1109/TVCG.2025.3540669}}

@article{3dcitymodel,
author = {Chaturvedi, Kanishk and Kolbe, Thomas},
year = {2016},
month = {10},
pages = {31-38},
title = {INTEGRATING DYNAMIC DATA AND SENSORS WITH SEMANTIC 3D CITY MODELS IN THE CONTEXT OF SMART CITIES},
volume = {IV-2/W1},
journal = {ISPRS Annals of the Photogrammetry, Remote Sensing and Spatial Information Sciences},
doi = {10.5194/isprs-annals-IV-2-W1-31-2016}
}

@article{iqbal2025exploring,
  title={Exploring the 15-minutes city concept: Global challenges and opportunities in diverse urban contexts},
  author={Iqbal, Asifa and Nazir, Humaira and Qazi, Ammad Waheed},
  journal={Urban Science},
  volume={9},
  number={7},
  pages={252},
  year={2025},
  publisher={MDPI}
}

@inproceedings{pan2024glomap,
    author={Pan, Linfei and Barath, Daniel and Pollefeys, Marc and Sch\"{o}nberger, Johannes Lutz},
    title={{Global Structure-from-Motion Revisited}},
    booktitle={European Conference on Computer Vision (ECCV)},
    year={2024},
}

@inproceedings{agarwal2011rome,
  title={Building Rome in a Day},
  author={Agarwal, Sameer and Furukawa, Yasutaka and Snavely, Noah and others},
  booktitle={ICCV},
  year={2011}
}

@inproceedings{schoenberger2016sfm,
  title={Structure-from-Motion Revisited},
  author={Sch{\"o}nberger, Johannes L. and Frahm, Jan-Michael},
  booktitle={CVPR},
  year={2016}
}

@inproceedings{orsingher2022revisiting,
  title={Revisiting patchmatch multi-view stereo for urban 3d reconstruction},
  author={Orsingher, Marco and Zani, Paolo and Medici, Paolo and Bertozzi, Massimo},
  booktitle={2022 IEEE Intelligent Vehicles Symposium (IV)},
  pages={190--196},
  year={2022},
  organization={IEEE}
}

@article{barron2022mipnerf360,
    title={Mip-NeRF 360: Unbounded Anti-Aliased Neural Radiance Fields},
    author={Jonathan T. Barron and Ben Mildenhall and 
            Dor Verbin and Pratul P. Srinivasan and Peter Hedman},
    journal={CVPR},
    year={2022}
}

@article{Held2025MeshSplatting,
  title = {MeshSplatting: Differentiable Rendering with Opaque Meshes},
  author = {Held, Jan and Son, Sanghyun and Vandeghen, Renaud and Rebain, Daniel and Gadelha, Matheus and Zhou, Yi and Cioppa, Anthony and Lin, Ming C. and Van Droogenbroeck, Marc and Tagliasacchi, Andrea},
  journal = {arXiv},
  year = {2025},
}

@misc{mai2025radiancemeshesvolumetricreconstruction,
      title={Radiance Meshes for Volumetric Reconstruction}, 
      author={Alexander Mai and Trevor Hedstrom and George Kopanas and Janne Kontkanen and Falko Kuester and Jonathan T. Barron},
      year={2025},
      eprint={2512.04076},
      archivePrefix={arXiv},
      primaryClass={cs.GR},
      url={https://arxiv.org/abs/2512.04076}, 
}

@article{guedon2025milo,
  title={Milo: Mesh-in-the-loop gaussian splatting for detailed and efficient surface reconstruction},
  author={Gu{\'e}don, Antoine and Gomez, Diego and Maruani, Nissim and Gong, Bingchen and Drettakis, George and Ovsjanikov, Maks},
  journal={ACM Transactions on Graphics (TOG)},
  volume={44},
  number={6},
  pages={1--15},
  year={2025},
  publisher={ACM New York, NY, USA}
}

@misc{wang2025mogeunlockingaccuratemonocular,
      title={MoGe: Unlocking Accurate Monocular Geometry Estimation for Open-Domain Images with Optimal Training Supervision}, 
      author={Ruicheng Wang and Sicheng Xu and Cassie Dai and Jianfeng Xiang and Yu Deng and Xin Tong and Jiaolong Yang},
      year={2025},
      eprint={2410.19115},
      archivePrefix={arXiv},
      primaryClass={cs.CV},
      url={https://arxiv.org/abs/2410.19115}, 
}
}

\clearpage
\setcounter{page}{1}

\newcommand{\supptitle}{City-Mesh3R: Simulation-Ready City-Scale 3D Mesh Reconstruction \\ from Multi-View Images}
\newcommand{\suppauthors}{
Sayan Paul \orcidlink{0000-0001-9885-233X}, Sourav Ghosh, Siddharth Katageri,
 Soumyadip Maity, Sanjana Sinha, Brojeshwar Bhowmick
}
\newcommand{\suppemails}{
{\small \{p.sayan, g.sourav10, siddharth.katageri, soumyadip.maity, sanjana.sinha, b.bhowmick\} $@$ tcs.com}
}
\newcommand{\suppaffil}{
{\small \textbf{Visual Computing \& Embodied AI Lab, TCS Research, India}} \\
}

\newcommand{\mycustomsupptitle}{
  \twocolumn[
    \centering
    \Large
    \textbf{\supptitle}\par
    \vspace{0.5em}
    Supplementary Material\par
    \vspace{0.75em}
    \normalsize
    \suppauthors\par
    \vspace{0.25em}
    \suppemails\par
    \suppaffil\par
    \vspace{1em}
  ]
}

\mycustomsupptitle

\section{Area partitioning of large-scale sparse SfM}
\label{sec:supp_details_area_ranking}

\subsection{Support-plane parameterization and regular partition construction}
\label{sec:supp_area_partitioning}

This subsection provides the implementation details omitted from the main paper (Sec.~\ref{sec:area_decomp_ranking}) for constructing the spatial partitions used before camera selection.

\paragraph{Dominant-support-plane parameterization.}
Let $\mathcal{P}=\{P_p\}$ denote the sparse 3D points reconstructed by COLMAP. We estimate a dominant support plane from $\mathcal{P}$ via RANSAC plane fitting, yielding parameters $(\mathbf{n},d)$ such that
\begin{equation}
\mathbf{n}^{\top}\mathbf{x}+d=0,
\end{equation}
where $\mathbf{n}\in\mathbb{R}^3$ is a unit normal and $d\in\mathbb{R}$ is the offset. This introduces only a weak geometric prior: unlike Manhattan-world alignment, we assume only one dominant support plane.

From the plane inliers, we construct a local planar frame $(\mathbf{o},\mathbf{u},\mathbf{v})$, where $\mathbf{o}$ is a point on the plane and $\mathbf{u},\mathbf{v}$ are orthonormal in-plane basis vectors. Each sparse point $P_p$ is mapped to planar coordinates by
\begin{equation}
\mathbf{r}_p=P_p-\mathbf{o},
\qquad
u_p=\mathbf{r}_p^\top\mathbf{u},
\qquad
v_p=\mathbf{r}_p^\top\mathbf{v}.
\end{equation}

\paragraph{In-plane orientation refinement.}
After fixing the plane normal, the in-plane basis remains ambiguous up to a rotation around $\mathbf{n}$. To reduce this arbitrariness, we choose the in-plane rotation that minimizes the area of the axis-aligned bounding box of the projected sparse points. Let $(u_p^{(\phi)},v_p^{(\phi)})$ denote the planar coordinates after rotation by angle $\phi$. We choose
\begin{equation}
\phi^\star
=
\arg\min_{\phi}
\left(
\max_p u_p^{(\phi)}-\min_p u_p^{(\phi)}
\right)
\left(
\max_p v_p^{(\phi)}-\min_p v_p^{(\phi)}
\right).
\end{equation}

\paragraph{Regular grid partitioning with inflated windows.}
Using the rotated planar coordinates, we define a regular $R\times C$ grid over the support-plane extent. Let
\begin{equation}
[u_{\min},u_{\max}] \times [v_{\min},v_{\max}]
\end{equation}
be the planar bounding box. The base grid boundaries are
\begin{equation}
u_c = u_{\min} + \frac{c}{C}(u_{\max}-u_{\min}),
\qquad
v_r = v_{\min} + \frac{r}{R}(v_{\max}-v_{\min}),
\end{equation}
for $c=0,\dots,C$ and $r=0,\dots,R$.

For each grid cell $(r,c)$ with base rectangle $[u_c,u_{c+1}] \times [v_r,v_{r+1}]$, we enlarge its support around the cell center
\begin{equation}
u^{(m)}_{r,c}=\frac{u_c+u_{c+1}}{2},
\qquad
v^{(m)}_{r,c}=\frac{v_r+v_{r+1}}{2}.
\end{equation}
Given inflation factors $\alpha_u$ and $\alpha_v$, the enlarged side lengths are
\begin{equation}
\Delta u'=(1+\alpha_u)(u_{c+1}-u_c),
\qquad
\Delta v'=(1+\alpha_v)(v_{r+1}-v_r).
\end{equation}
The inflated window becomes
\begin{equation}
\hat{u}_0=u^{(m)}_{r,c}-\frac{\Delta u'}{2},
\qquad
\hat{u}_1=u^{(m)}_{r,c}+\frac{\Delta u'}{2},
\end{equation}
\begin{equation}
\hat{v}_0=v^{(m)}_{r,c}-\frac{\Delta v'}{2},
\qquad
\hat{v}_1=v^{(m)}_{r,c}+\frac{\Delta v'}{2},
\end{equation}
followed by clipping to the global planar extent. The point set assigned to partition $(r,c)$ is
\begin{equation}
\mathcal{P}_{r,c}
=
\left\{
P_p \in \mathcal{P}
\;\middle|\;
\hat{u}_0 \le u_p \le \hat{u}_1,\;
\hat{v}_0 \le v_p \le \hat{v}_1
\right\}.
\end{equation}

These partition point sets are used by the camera-ranking procedure described next.

\subsection{Per-partition geometry-aware camera ranking}
\label{sec:supp_camera_ranking}

This subsection gives the full formulation of the partition-wise camera ranking summarized in the main paper (Sec.~\ref{sec:area_decomp_ranking}).

For each non-empty partition $(r,c)$, let $\mathcal{P}_{r,c}$ denote the partition point set. We first form the candidate camera set as the union of all cameras that observe at least one partition point:
\begin{equation}
\mathcal{I}^{\mathrm{cand}}_{r,c}
=
\bigcup_{p\in\mathcal{P}_{r,c}} \mathcal{I}(p).
\end{equation}
If $\mathcal{P}_{r,c}=\emptyset$, or if no sufficiently co-visible camera pairs exist within the partition, we return an empty ranked set.

\paragraph{Admissible camera pairs.}
For each unordered camera pair $(i,j)$, let $m_{ij}$ be the number of points in $\mathcal{P}_{r,c}$ observed by both cameras, and let $N_{r,c}=|\mathcal{P}_{r,c}|$. We retain only admissible pairs whose co-visibility ratio exceeds a threshold:
\begin{equation}
\rho_{ij} = \frac{m_{ij}}{N_{r,c}},
\qquad
\mathcal{A} = \{(i,j)\mid \rho_{ij}\ge \tau_{\mathrm{cov}}\}.
\end{equation}

\paragraph{Pair prior.}
For each admissible pair $(i,j)\in\mathcal{A}$, we define a prior that captures viewpoint complementarity and local support. Let $\theta_{ij}$ denote the relative rotation angle between cameras $i$ and $j$. We favor moderate relative orientation using
\begin{equation}
P_{\mathrm{rot}}(i,j)
=
\exp\!\left(
-\frac{(\theta_{ij}-\mu_{\mathrm{rot}})^2}{2\sigma_{\mathrm{rot}}^2}
\right).
\end{equation}
We also map the pairwise co-visibility count to a normalized overlap term
\begin{equation}
P_{\mathrm{overlap}}(i,j)
=
\frac{\log(1+m_{ij})}{\log(1+m_{\max})},
\qquad
m_{\max}=\max_{(i,j)\in\mathcal{A}} m_{ij}.
\end{equation}
The resulting pair prior is
\begin{equation}
\operatorname{prior}_{ij}
=
P_{\mathrm{rot}}(i,j)^{w_{\mathrm{rot}}}
\cdot
P_{\mathrm{overlap}}(i,j)^{w_{\mathrm{overlap}}}.
\end{equation}

\paragraph{Point-conditioned pair scoring.}
For each point $p\in\mathcal{P}_{r,c}$, we consider only admissible pairs that observe it:
\begin{equation}
\mathcal{A}(p)=\{(i,j)\in\mathcal{A}\mid i,j\in\mathcal{I}(p)\}.
\end{equation}
If $\mathcal{A}(p)=\emptyset$, the point contributes no vote.

Let $C_i$ and $C_j$ be the camera centers, and let $P_p$ be the 3D point. Using the normalized viewing rays from $C_i$ and $C_j$ toward $P_p$, we compute the triangulation angle $\alpha_{ij}(p)$ and define
\begin{equation}
T_{ij}(p)
=
\exp\!\left(
-\frac{(\alpha_{ij}(p)-\mu_{\mathrm{triang}})^2}{2\sigma_{\mathrm{triang}}^2}
\right).
\end{equation}

We also project $P_p$ into images $i$ and $j$, compute the normalized radial distances $r_i(p)$ and $r_j(p)$ from the principal points, and define centrality weights
\begin{equation}
C_i(p)
=
\exp\!\left(
-\frac{r_i(p)^2}{2\sigma_{\mathrm{cent}}^2}
\right),
\qquad
C_j(p)
=
\exp\!\left(
-\frac{r_j(p)^2}{2\sigma_{\mathrm{cent}}^2}
\right).
\end{equation}
These are combined symmetrically as
\begin{equation}
C_{ij}(p)=\sqrt{C_i(p)\,C_j(p)}.
\end{equation}

The final point-pair score is
\begin{equation}
s_{p,ij}
=
\operatorname{prior}_{ij}
\cdot
T_{ij}(p)^{w_{\mathrm{triang}}}
\cdot
C_{ij}(p)^{w_{\mathrm{cent}}}.
\end{equation}

\begin{figure}[h]
    \centering
    \includegraphics[width=0.5\textwidth]{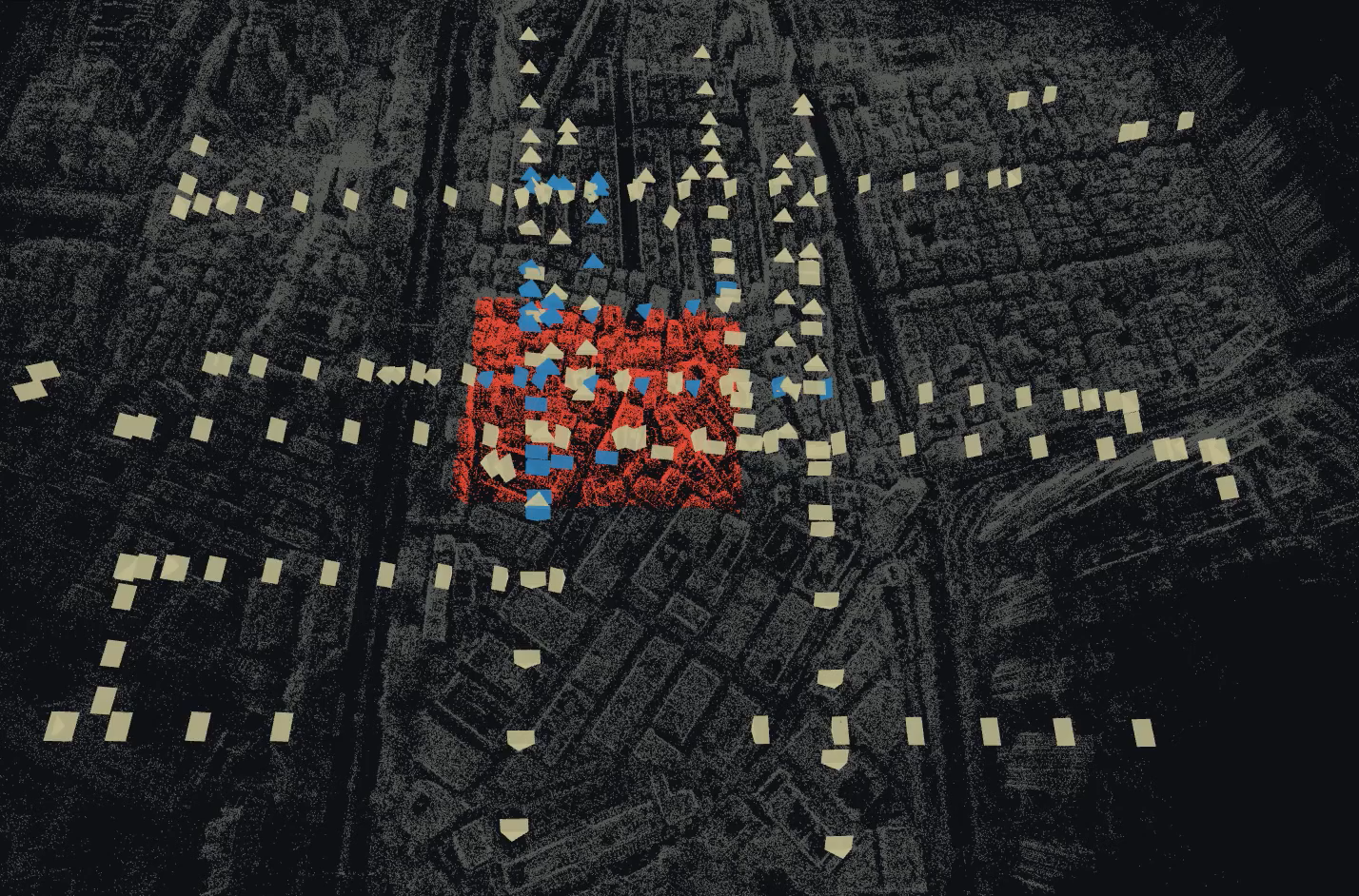}
    \caption{\textbf{Area Partitioning Visualizer:} The selected partition's sparse points are highlighted by orange and all other points are shown in gray. All the candidate cameras of a partition are denoted by yellow and the top-M ranked cameras are denoted by blue.}
    \label{fig:Area_Partitioning_Vis_01}
\end{figure}

\paragraph{Camera score aggregation.}
Initialize $S_i=0$ for all $i\in\mathcal{I}^{\mathrm{cand}}_{r,c}$. To avoid over-counting many near-equivalent pairs for one point, we retain only the top-$K$ scoring pairs $\tilde{\mathcal{A}}(p)\subseteq\mathcal{A}(p)$ and normalize their scores:
\begin{equation}
w_{p,ij}
=
\frac{s_{p,ij}}
{\sum_{(i,j)\in\tilde{\mathcal{A}}(p)} s_{p,ij} + \varepsilon}.
\end{equation}
Each selected pair casts a unit-normalized vote split equally between its two cameras:
\begin{equation}
S_i \leftarrow S_i + \tfrac{1}{2}w_{p,ij},
\qquad
S_j \leftarrow S_j + \tfrac{1}{2}w_{p,ij},
\qquad
(i,j)\in\tilde{\mathcal{A}}(p).
\end{equation}
Aggregating these votes over all points yields a partition-specific score for every candidate camera. We then sort cameras by $S_i$ and retain the top $M$:
\begin{equation}
\mathcal{I}^{\mathrm{top}}_{r,c}
=
\operatorname{TopM}\!\left(\{S_i\}_{i\in\mathcal{I}^{\mathrm{cand}}_{r,c}}, M\right).
\end{equation}

\section{Dense Reconstruction and Surface Initialization}
\label{sec:supp_dense_init}

This section provides the detailed formulation for the dense reconstruction and surface initialization module summarized in \Cref{sec:dense_init}.

For each partitioned area, we start from its sparse SfM reconstruction and retain the top-$M$ cameras from the ranking stage. Their intrinsics and poses, $\{K_n,R_n,t_n\}_{n=1}^{M}$, are kept fixed throughout. We then densify the partition using the depth predictions and pixel correspondences produced by MASt3R. Since MASt3R inference has already been run on image pairs in the original similarity graph, we reuse the cached outputs whenever possible; only pairs newly induced by the final partitioning require fresh inference.

Let $\mathcal{V}$ denote the selected views and $\mathcal{E}$ the set of view pairs for which MASt3R correspondences are available after this cache-or-infer step. For each view $n\in\mathcal{V}$, let $D_n$ denote the predicted depth map, and let $\mathcal{M}^{n,m}$ be the set of matched pixels between views $n$ and $m$. For a correspondence $u_c^n \leftrightarrow u_c^m$, we define the associated world-space dense points by back-projecting each pixel using the fixed SfM camera:
\[
X_c^n = \Pi_n^{-1}(u_c^n, D_n(u_c^n)),
\qquad
X_c^m = \Pi_m^{-1}(u_c^m, D_m(u_c^m)).
\]
Thus, $X_c^n$ is the 3D point induced by pixel $u_c^n$ from the predicted depth of view $n$, expressed in the common SfM world frame. Equivalently, the collection $\{X_c^n\}_c$ can be interpreted as the dense pointmap of view $n$ in world coordinates, consistent with the pointmap representation used in MASt3R.

Although the cameras are fixed, these per-view dense reconstructions are not necessarily globally consistent: they may still suffer from residual scale drift across views and local depth noise. To address this, we adopt a two-stage alignment strategy inspired by MASt3R-SfM, but specialized to our setting by optimizing only scene geometry while keeping all camera parameters frozen.

\paragraph{Coarse alignment of Depth Maps: }
We first assign each view a global scale $s_n>0$ and rescale its dense points as $\widetilde{X}_c^n = s_n X_c^n$. The scales are estimated by minimizing the discrepancy between matched 3D points:
\begin{equation}
\label{eq:supp_coarse_scale}
\{s_n^*\}
=
\arg\min_{\{s_n\}}
\sum_{(n,m)\in\mathcal{E}}
\sum_{c\in\mathcal{M}^{n,m}}
q_c \,
\| s_n X_c^n - s_m X_c^m \|_2^{\lambda_1},
\end{equation}
where $q_c$ denotes the confidence of match $c$, and $0<\lambda_1\leq 1$ defines a robust penalty. To remove the gauge ambiguity, we normalize the estimated scales as
\[
s_n = s_n' / \min_k s_k'.
\]
This stage allows each view to expand or contract globally so that corresponding dense points become metrically compatible across overlapping views.

\paragraph{Depth Refinement: }
A single scale per view is insufficient to correct local depth errors. We therefore further refine the geometry by directly optimizing the dense 3D points while keeping the cameras fixed. Let $\widehat{X}_c^n$ denote the refined 3D points. We optimize them by minimizing the bidirectional reprojection inconsistency:
\begin{equation}
\label{eq:dense_refine}
\begin{aligned}
\{\widehat{X}_c^n\}
=
\arg\min_{\{\widehat{X}_c^n\}}
\sum_{(n,m)\in\mathcal{E}}
\sum_{c\in\mathcal{M}^{n,m}}
q_c \big[
& \rho(u_c^n - \Pi_n(\widehat{X}_c^m)) \\
& + \rho(u_c^m - \Pi_m(\widehat{X}_c^n))
\big],
\end{aligned}
\end{equation}
where $\Pi_n(\cdot)$ is the projection operator of the fixed camera $n$, and $\rho(x)=\|x\|_2^{\lambda_2}$ with $0<\lambda_2\leq 1$ is a robust reprojection penalty. This objective is analogous to bundle adjustment, except that only the scene geometry is optimized. The coarse stage removes dominant inter-view scale mismatch, while the refinement stage suppresses local geometric noise by enforcing multi-view reprojection consistency.

Both stages are optimized with Adam for a fixed number of iterations, yielding aligned dense depth maps for the partition.

\paragraph{Volumetric fusion and watertight initialization.}
The aligned depth maps are fused using TSDF integration, which aggregates multi-view evidence and regularizes local inconsistencies through volumetric averaging. From the TSDF grid, we extract a fused global point cloud with normals. These oriented points are then passed to Screened Poisson Surface Reconstruction to obtain an initial watertight mesh, which is used to initialize the subsequent differentiable-rendering-based mesh refinement stage.

In summary, this module converts fixed sparse-SfM cameras and mostly cached MASt3R predictions into a dense, globally aligned, and denoised surface initialization, without re-estimating camera poses.

\begin{figure*}[ht!]
    \centering
    \includegraphics[width=\textwidth]{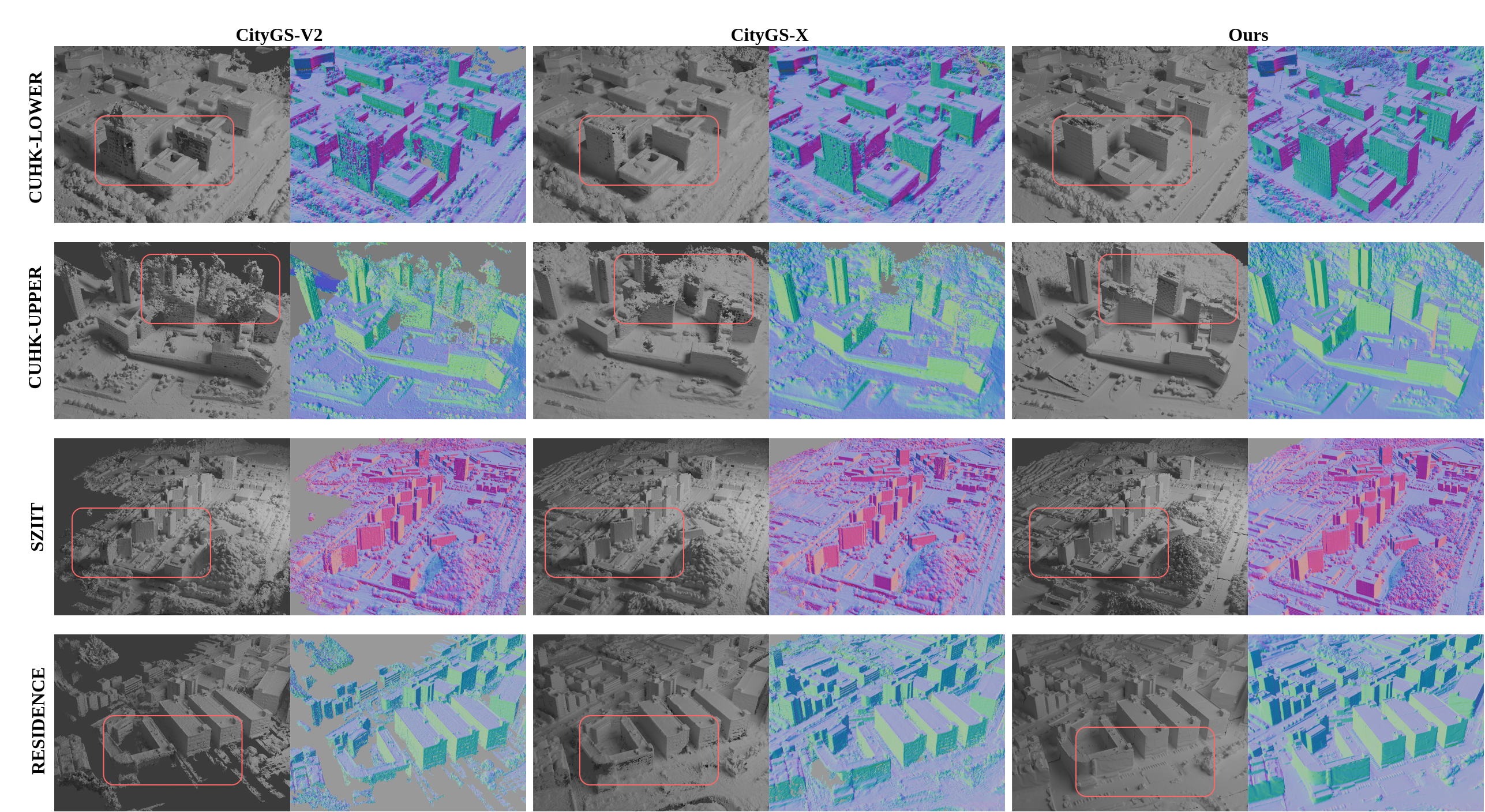}
    \caption{Extra Qualitative Comparison Results with recent city-scale surface reconstruction methods (CityGS-v2 and CityGS-X) on GauU-Scene dataset (CUHK-LOWER, CUHK-UPPER, SZIIT) and UrbanScene3D dataset (Residence). Go to \cref{fig:Qual_main} for Main Results.}
    \label{fig:Qual_supply}
\end{figure*}
\section{Differentiable Rendering based Mesh Refinement}
\label{sec:supp_mesh_refinement_details}

This section provides the full formulation and implementation details omitted from \cref{sec:mesh_refinement}.

\subsection{Rendering Objective}
\label{subsec:supp_objective}

At iteration \(k\), the current mesh is \(M^k=(V^k,F^k)\). Given calibrated cameras \(\{K_j,T_{wc,j}\}_{j=1}^{N}\), we optimize
\begin{equation}
\Phi(M^k)=
\sum_{j=1}^{N}
\left(
\lambda_n\,\mathcal L_n^{(j)}(M^k)+
\lambda_s\,\mathcal L_{\mathrm{sil}}^{(j)}(M^k)
\right)+
\mathcal R(M^k),
\end{equation}
where \(\mathcal R\) is a lightweight mesh regularizer, e.g.\ Laplacian smoothing.

For view \(j\), let \(S_j(u)\in[0,1]\) be the target silhouette and \(\hat S_j(u;M^k)\) the rendered silhouette. The silhouette loss is
\begin{equation}
\mathcal L_{\mathrm{sil}}^{(j)}(M^k)=
\frac{1}{|\Omega_{\mathrm{img}}|}
\sum_{u\in\Omega_{\mathrm{img}}}
\big(\hat S_j(u;M^k)-S_j(u)\big)^2.
\end{equation}
Let \(\mathcal N_j(u)\in\mathbb S^2\) be the predicted unit normal map in camera coordinates and \(\hat{\mathcal N}_j(u;M^k)\) the rendered unit normal map. On the foreground support \(\Omega_j=\{u\mid S_j(u)=1\}\), the normal-map loss is
\begin{equation}
\mathcal L_n^{(j)}(M^k)=
\frac{1}{|\Omega_j|}
\sum_{u\in\Omega_j}
\left(
1-\hat{\mathcal N}_j(u;M^k)^\top \mathcal N_j(u)
\right).
\end{equation}

Gradients of \(\Phi\) are backpropagated through the renderer to vertex positions, which are updated using the IsotropicAdam optimizer of Continuous Remeshing. As in the original framework, the optimizer also provides a relative vertex velocity \(\nu^k(i)\in[0,1]\), which measures the normalized update magnitude at vertex \(i\) and is used as a local indicator of stabilization.

\subsection{Curvature-Guided Reference Field}
\label{subsec:supp_curvature_field}

The original controller tends to distribute resolution relatively uniformly, which is inefficient for large scenes: flat regions consume unnecessary triangles, while detailed regions remain under-sampled. We instead decouple \emph{where} fine resolution is ultimately needed from \emph{when} that refinement is allowed.

For each view \(j\), let \(\mathbf n^{(j)}(u,v)\in\mathbb S^2\) be the unit normal field, with image-space Jacobian
\begin{equation}
N^{(j)}(u,v)=
\begin{bmatrix}
\partial_u \mathbf n^{(j)}(u,v) & \partial_v \mathbf n^{(j)}(u,v)
\end{bmatrix}
\in\mathbb R^{3\times 2}.
\end{equation}
We define
\begin{equation}
M^{(j)}(u,v)=\big(N^{(j)}(u,v)\big)^\top N^{(j)}(u,v)\in\mathbb R^{2\times 2},
\end{equation}
and the worst-case local normal-rotation rate
\begin{equation}
s^{(j)}(u,v)=
\sqrt{\lambda_{\max}\!\left(M^{(j)}(u,v)\right)}=
\|N^{(j)}(u,v)\|_2.
\end{equation}
Given a normal-rotation tolerance \(\theta_0\) (radians), the target projected edge length in pixels is
\begin{equation}
p_{\mathrm{tgt}}^{(j)}(u,v)=
\mathrm{clip}\!\left(\frac{\theta_0}{s^{(j)}(u,v)},\,p_{\min},\,p_{\max}\right).
\end{equation}
Hence, regions with rapid normal variation receive smaller projected target edges, while smooth regions permit coarser sampling.

We then lift this per-view pixel-domain target to the mesh surface. For a vertex \(i\) visible in view \(j\), let \(\pi^{(j)}(\mathbf v_i)\) be its projection and \(z_i^{(j)}\) its camera-space depth. Sampling the pixel target gives \(p_{ij}=p_{\mathrm{tgt}}^{(j)}(\pi^{(j)}(\mathbf v_i))\), which is converted to world units as
\begin{equation}
L_{ij}=\frac{z_i^{(j)}}{f^{(j)}}\,p_{ij},
\end{equation}
where \(f^{(j)}\) is the focal length in pixels. Pooling across visible views \(\mathcal V_i\) yields the per-vertex curvature-guided baseline
\begin{equation}
L_{\mathrm{ref\text{-}curv}}(i)=\operatorname*{median}_{j\in\mathcal V_i} L_{ij}.
\end{equation}

In practice, remeshing acts on edge targets derived from the vertex field by endpoint averaging. This lets us maintain the reference field on vertices while applying split/collapse decisions on edges.

\subsection{Speed-Aware Slack Schedule}
\label{subsec:supp_slack}

To control when refinement is released, we introduce a nonnegative slack \(\zeta^k(i)\) on top of the curvature-guided baseline. The slack is initialized as
\begin{equation}
\zeta^0(i)\leftarrow
\max\!\big(0,\;L_{\mathrm{curr}}^0(i)-L_{\mathrm{ref\text{-}curv}}^0(i)\big),
\end{equation}
where \(L_{\mathrm{curr}}^0(i)\) denotes the current local edge scale at initialization.

The slack is then updated using the optimizer's relative vertex velocity:
\begin{equation}
\zeta^{k}(i)\leftarrow
\max\!\Big(
0,\;
\zeta^{k-1}(i)+g\big(\nu^k(i)-\nu_{\mathrm{ref}}\big)
\Big),
\end{equation}
where \(g>0\) is a gain and \(\nu_{\mathrm{ref}}\) is a reference speed. The resulting per-vertex reference edge length is
\begin{equation}
L_{\mathrm{ref}}^{k}(i)=
\mathrm{clip}\!\Big(
L_{\mathrm{ref\text{-}curv}}^{k}(i)+\zeta^k(i),\,
L_{\min},\,
L_{\max}
\Big).
\end{equation}
The corresponding per-edge reference length is obtained from the two incident vertex values. Intuitively, curvature specifies where fine resolution is eventually required, while velocity controls when the local mesh is allowed to approach that target.

\subsection{Automatic Estimation of Global Length Bounds}
\label{subsec:supp_bounds}

The global bounds \(L_{\min}\) and \(L_{\max}\) are estimated automatically from camera intrinsics and a robust scene-depth statistic. Let \(f_x=K_{00}\), \(f_y=K_{11}\), and \(f_{\mathrm{iso}}=\sqrt{f_xf_y}\). Flatten all valid rendered depths of the initial mesh and let \(z_p\) be a robust percentile. Then
\begin{equation}
\mathrm{wlpp}=\frac{z_p}{f_{\mathrm{iso}}},\qquad
L_{\min}=\mathrm{wlpp}.
\end{equation}
If \(\mathcal E\) denotes the set of unique mesh edges and
\begin{equation}
L_{\mathrm{med}}=\operatorname{median}_{(i,m)\in\mathcal E}\|\mathbf v_i-\mathbf v_m\|_2,
\end{equation}
we set
\begin{equation}
L_{\max}=
\max\big(k_1L_{\mathrm{med}},\;k_2L_{\min}\big),
\end{equation}
with \(k_1,k_2>1\). This prevents both sub-pixel triangles and overly coarse edges.

\begin{figure}[h]
    \centering
    \includegraphics[width=0.5\textwidth]{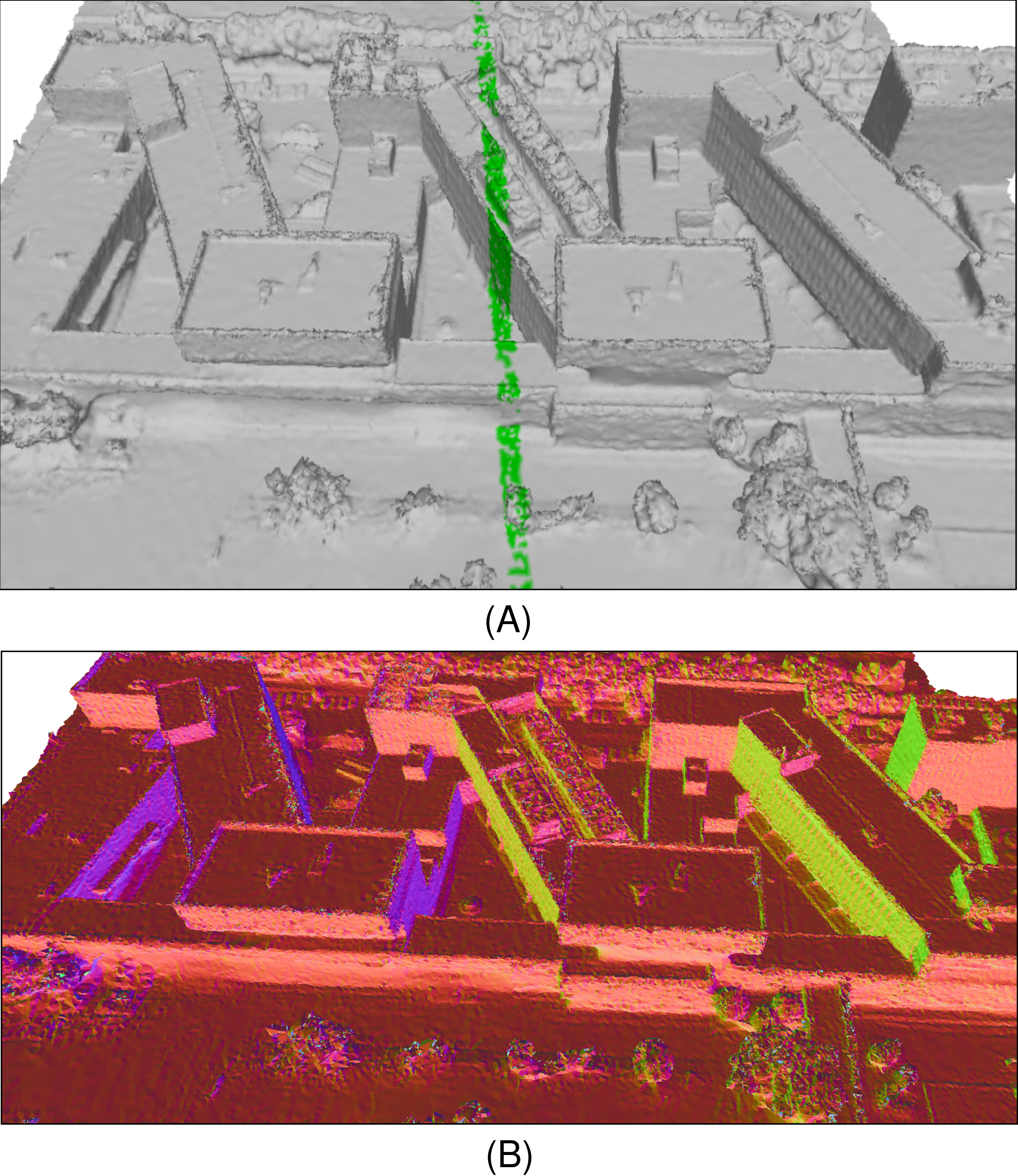}
    \caption{\textbf{Mesh Stitching Result:}
    (A) Two adjacent partition meshes with seam-region points highlighted in green. (B) Surface normals of the final stitched mesh. Our seam stitching yields a topologically consistent and geometrically smooth merge, preserving clean boundaries and normal continuity across partitions.}
    \label{fig:MeshMerge}
\end{figure}

\subsection{Remeshing Rules}
\label{subsec:supp_remeshing}

After each vertex update, we invoke the remesher under the current reference field \(L_{\mathrm{ref}}^k\). As in Continuous Remeshing, remeshing is embedded directly into the optimization loop and consists of local \textbf{collapse}, \textbf{split}, and \textbf{flip} operations. Let \(\ell(e)\) denote the current length of edge \(e\). Edges are collapsed when
\begin{equation}
\ell(e) < \tau_c\,L_{\mathrm{ref}}^k(e),
\end{equation}
split when
\begin{equation}
\ell(e) > \tau_s\,L_{\mathrm{ref}}^k(e),
\end{equation}
and flipped when doing so improves local triangle quality and valence regularity, subject to the same geometric validity checks as in the original framework. Thus, the remesher keeps local edge lengths near the target field while preserving a well-shaped triangulation.

The overall refinement alternates render \(\rightarrow\) compare \(\rightarrow\) update \(\rightarrow\) remesh, and the final mesh is selected as the best snapshot \(M^\star=\arg\min_k \Phi(M^k)\).

\section{Mesh Quality Metrics}
\label{mesh_quality_metrics_defn}
We assess mesh quality using the following metrics, which quantify complementary aspects of geometric validity, topological consistency, and structural regularity.

\begin{table*}[ht]
\centering
\setlength{\tabcolsep}{6pt}
\renewcommand{\arraystretch}{1.25}
\begin{tabular}{l c c c c}
\toprule
\textbf{Metric} &
\textbf{Ours} &
\textbf{MiLo} &
\textbf{MeshSplatting} &
\textbf{RadianceMesh} \\
&
\scriptsize 488K V · 973K F &
\scriptsize 1384K V · 2771K F &
\scriptsize 1190K V · 3503K F &
\scriptsize 483K V · 1101K F \\
\midrule
Aspect Ratio (AR) $\downarrow$
  & \textcolor{ForestGreen}{\textbf{3.025}} & 15.652 & 5.224 & \textcolor{blue}{3.147} \\

Angle bad ratio (ANG) $\downarrow$
  & \textcolor{ForestGreen}{\textbf{1.92\%}} & 28.83\% & 15.25\% & \textcolor{blue}{2.93\%} \\

Degenerate Triangle Ratio (DTR) $\downarrow$
  & \textcolor{ForestGreen}{\textbf{0.0000}} & 0.0003 & \textcolor{ForestGreen}{\textbf{0.0000}} & \textcolor{ForestGreen}{\textbf{0.0000}} \\

Non-Manifold Edge Ratio (NME) $\downarrow$
  & \textcolor{blue}{0.0039\%} & \textcolor{ForestGreen}{\textbf{0.0000\%}} & 27.34\% & 1.28\% \\

Non-Manifold Vertex Ratio (NMV) $\downarrow$
  & \textcolor{blue}{0.0027\%} & \textcolor{ForestGreen}{\textbf{0.0014\%}} & 31.34\% & 6.26\% \\

Vertex Valence Deviation (VVD) $\downarrow$
  & \textcolor{ForestGreen}{\textbf{0.505}} & \textcolor{blue}{1.421} & 3.045 & 1.914 \\

Connected Components (CC) $\downarrow$
  & \textcolor{ForestGreen}{\textbf{9}} & 918 & 340 & \textcolor{blue}{232} \\

Interior Boundary Loops (IBL) $\downarrow$
  & \textcolor{ForestGreen}{\textbf{24}} & 61 & 6554 & \textcolor{blue}{58} \\
\bottomrule
\end{tabular}
\caption{Mesh Quality Metric Comparison across four reconstruction methods on the Garden scene of the MipNeRF360 dataset. Best values are highlighted in \textcolor{ForestGreen}{\textbf{green}} and bold, second-best in \textcolor{blue}{blue}. Lower is better for all metrics. For the metric definitions, please refer to \cref{mesh_quality_metrics_defn}.}
\label{tab:mesh_quality}
\end{table*}

\paragraph{Aspect Ratio (AR).}
For a triangular face $f \in \mathcal{F}$, let $h_f$ denote its longest edge length and $r_f$ its inradius. The face-wise aspect ratio is defined as
\begin{equation}
\mathrm{AR}(f) = \frac{h_f}{2r_f},
\end{equation}
and the mesh-level aspect ratio is given by
\begin{equation}
\mathrm{AR} = \frac{1}{|\mathcal{F}|} \sum_{f \in \mathcal{F}} \mathrm{AR}(f).
\end{equation}
Lower values indicate better-shaped triangles.

\paragraph{Angle Bad Ratio (ANG).}
Let $\theta_{\min}(f)$ and $\theta_{\max}(f)$ denote the minimum and maximum interior angles of face $f$, respectively. Given angular thresholds $\theta_{\mathrm{lo}}$ and $\theta_{\mathrm{hi}}$, the angle bad ratio is defined as
\begin{equation}
\mathrm{ANG} =
\frac{
\left|
\left\{
f \in \mathcal{F}
\;\middle|\;
\theta_{\min}(f) < \theta_{\mathrm{lo}}
\;\text{or}\;
\theta_{\max}(f) > \theta_{\mathrm{hi}}
\right\}
\right|
}{
|\mathcal{F}|
}.
\end{equation}
This metric measures the fraction of triangles with poor angular quality. Lower values are preferred.

\paragraph{Degenerate Triangle Ratio (DTR).}
Let $A(f)$ denote the area of face $f \in \mathcal{F}$, and let $\varepsilon_A > 0$ denote a small numerical threshold. The degenerate triangle ratio is defined as
\begin{equation}
\mathrm{DTR} =
\frac{
|\{ f \in \mathcal{F} \mid A(f) \le \varepsilon_A \}|
}{
|\mathcal{F}|
}.
\end{equation}
This metric quantifies the proportion of triangles that are degenerate or nearly degenerate. Lower values are better.

\paragraph{Non-Manifold Edge Ratio (NME).}
Let $\mathcal{E}$ denote the set of mesh edges, and let $\deg(e)$ denote the number of incident faces of edge $e \in \mathcal{E}$. The non-manifold edge ratio is defined as
\begin{equation}
\mathrm{NME} =
\frac{
|\{ e \in \mathcal{E} \mid \deg(e) > 2 \}|
}{
|\mathcal{E}|
}.
\end{equation}
Lower values indicate better topological consistency.

\paragraph{Non-Manifold Vertex Ratio (NMV).}
Let $\mathcal{V}$ denote the set of mesh vertices. The non-manifold vertex ratio is defined as
\begin{equation}
\mathrm{NMV} =
\frac{
|\{ v \in \mathcal{V} \mid v \text{ is non-manifold} \}|
}{
|\mathcal{V}|
}.
\end{equation}
Lower values indicate fewer vertex-level topological defects.

\paragraph{Vertex Valence Deviation (VVD).}
For each vertex $v \in \mathcal{V}$, let $\mathrm{val}(v)$ denote its valence, i.e., the number of one-ring neighboring vertices. We define vertex valence deviation as
\begin{equation}
\mathrm{VVD} =
\frac{1}{|\mathcal{V}|}
\sum_{v \in \mathcal{V}}
\left| \mathrm{val}(v) - 6 \right|.
\end{equation}
This metric measures deviation from regular triangular connectivity. Lower values indicate a more regular tessellation.

\paragraph{Connected Components (CC).}
Let $\mathcal{C}$ denote the set of connected components of the mesh under standard edge-based connectivity. We define
\begin{equation}
\mathrm{CC} = |\mathcal{C}|.
\end{equation}
Lower values indicate a more globally coherent mesh, with $\mathrm{CC}=1$ corresponding to a single connected surface.

\paragraph{Interior Boundary Loops (IBL).}
Let $\mathcal{C}_{\max}$ be the largest connected component of the mesh, and let $\mathcal{B}(\mathcal{C}_{\max})$ denote its boundary loops. For outdoor scene meshes, which are typically open surfaces, we treat the longest loop as the intended exterior boundary of the dominant surface, and define
\begin{equation}
\mathrm{IBL} =
\max\!\left(
|\mathcal{B}(\mathcal{C}_{\max})| - 1,\; 0
\right).
\end{equation}
IBL counts the remaining boundary loops after excluding this primary exterior boundary. Lower values indicate fewer residual open-boundary artifacts, such as unintended holes, tears, or missing patches. It does not capture handle- or genus-type topological holes that do not induce boundaries.

\subsection{Quantitative Results}

As shown in \Cref{tab:mesh_quality}, our method achieves the best overall mesh quality. It ranks first in aspect ratio ($3.025$), angle bad ratio ($1.92\%$), vertex valence deviation ($0.505$), connected components ($9$), and interior boundary loops ($24$), while also achieving zero degenerate triangles. This indicates better triangle quality, more regular tessellation, and markedly stronger global coherence than the baselines.

Although MiLo attains slightly lower non-manifold edge and vertex ratios, it remains highly fragmented ($918$ connected components, $61$ interior boundary loops). RadianceMesh also achieves zero degenerate triangles, but is substantially worse in connectivity and boundary quality ($232$ connected components, $58$ interior boundary loops). MeshSplatting further achieves zero degenerate triangles, yet suffers from severe topological defects, with very high non-manifold ratios ($27.34\%$ NME, $31.34\%$ NMV), $340$ connected components, and $6554$ interior boundary loops.

These results show that our method yields the most structurally clean and simulation-ready meshes among the compared approaches.

\section{Limitations}
Although our pipeline is highly scalable and produces clean, simulation-ready meshes, a few practical constraints remain.
Our method inherits any residual errors from upstream SfM or pretrained depth prediction models, and extremely texture-poor or reflective regions can still challenge local reconstruction quality. Seam merging is robust, but very fine details at partition boundaries may experience mild smoothing. Finally, as clusters are processed independently, the system currently does not leverage global structural priors such as repeated façades or urban symmetries. Overall, these limitations are localized and do not affect the method’s strong scalability, stability, and suitability for city-scale digital-twin applications.

\end{document}